# The quality of priority ratios estimation in relation to a selected prioritization procedure and consistency measure for a Pairwise Comparison Matrix

Paul Thaddeus KAZIBUDZKI

*Universite Internationale Jean-Paul II de Bafang*
*B.P. 213 Bafang, Cameroun*

Tel/Fax: +237.96.25.90.25

Email: emailpoczta@gmail.com

**Abstract**: An overview of current debates and contemporary research devoted to the modeling of decision making processes and their facilitation directs attention to the Analytic Hierarchy Process (AHP). At the core of the AHP are various prioritization procedures (PPs) and consistency measures (CMs) for a Pairwise Comparison Matrix (PCM) which, in a sense, reflects preferences of decision makers. Certainly, when judgments about these preferences are perfectly consistent (cardinally transitive), all PPs coincide and the quality of the priority ratios (PRs) estimation is exemplary. However, human judgments are very rarely consistent, thus the quality of PRs estimation may significantly vary. The scale of these variations depends on the applied PP and utilized CM for a PCM. This is why it is important to find out which PPs and which CMs for a PCM lead directly to an improvement of the PRs estimation accuracy. The main goal of this research is realized through the properly designed, coded and executed seminal and sophisticated simulation algorithms in *Wolfram Mathematica 8.0*. These research results convince that the embedded in the AHP and commonly applied, both genuine PP and CM for PCM may significantly deteriorate the quality of PRs estimation; however, solutions proposed in this paper can significantly improve the methodology.

## Introduction

It is agreed that the world is a complex system of interacting elements. It is obvious that human minds have not yet evolved to the point where they can clearly perceive relationships of this global system and solve crucial issues associated with them. In order to deal with complex and fuzzy social, economic, and political issues, people must be supported and guided on their way to order priorities, to agree that one goal out-weighs another from a perspective of certain established criterion, to make tradeoffs in order to be able to serve the greatest common interest (Caballero, Romero & Ruiz 2016; García-Melón et al. 2016).

Obviously, intuition cannot be trusted, although many commonly do so, attempting to devise solutions for complex problems which demand reliable answers. Overwhelming scientific evidence indicates that the unaided human brain is simply not capable of simultaneous analysis of many different competing factors and then synthesizing the results for the purpose of rational decision. It is presumably the principal reason why scientists continuously deal with explanations and modeling of decisional problems in a way to make them widely comprehensible. That is why many supportive methodologies have been elaborated in order to make the decision making process easier, more credible and sometimes even possible. Indeed, numerous psychological experiments (Martin 1973), including the well-known Miller study (Miller 1956) put forth the notion that humans are not capable of dealing accurately with more than about seven (±2) things at a time (the human brain is limited in its short term memory capacity, its discrimination ability and its bandwidth of perception).

## Principles of the analytic thinking process

Humans learn about anything by two means. The first involves examining and studying some phenomenon from the perspective of its various properties, and then synthesizing findings and drawing conclusions. The second entails studying some phenomenon in relation to other similar phenomena and relating them by making comparisons (Saaty 2008). The latter method leads directly to the essence of the matter i.e. judgments regarding the phenomenon. Judgments can be relative and absolute. An absolute judgment is the relation between a single stimulus and some information held in short or long term memory. A relative judgment, on the other hand, can be defined as the identification of some relation between two stimuli both present to the observer (Blumenthal 1977). It is said that humans can make much better relative judgments than absolute ones (Saaty 2000). It is probably so because humans have better ability to discriminate between the members of a pair, than compare one thing against some recollection from long term memory.

For detailed knowledge, the mind structures complex reality into its constituent parts, and these in turn into their elements. The number of parts usually ranges between five and nine. By breaking down reality into homogenous clusters and subdividing those into smaller ones, humans can integrate large amounts of information into the structure of a problem and form a more comprehensive picture of the whole system. Abstractly, this process entails the decomposition of a system into a hierarchy which is a model of a complex reality. Thus, a hierarchy constitutes a structure of multiple levels where the first level is the objective followed successively by levels of factors, criteria, sub-criteria, and so on down to a bottom level of alternatives. The goal of this hierarchy is to evaluate the influence of higher level elements on those of a lower level or alternatively the contribution of elements in the lower level to the importance or fulfillment of the elements in the level above. In this context the latter elements serve as criteria and are called properties.

Generally, a hierarchy can be functional or structural. The latter closely relates to the way a human brain analyzes complexity by breaking down the objects perceived by the senses into clusters and sub-clusters, and so on. Thus, in structural hierarchies, complex systems are structured into their constituent parts in descending order according to their structural properties. In contrast, in functional hierarchies complex systems are decomposed into their constituent parts in accordance to their essential relationships.

A large number of hierarchies in application are available in the literature (Saaty 1993). Supposedly, the hierarchical classification is the most powerful method applied by the human mind during intellectual reasoning and ordering of information and/or observations. Thus, we may agree that an efficient and effective multiple criteria decision making process should encompass the following steps:
– transpose the problem into a hierarchy;
– derive judgments that reflect ideas and feelings or emotions;
– represent these judgments with meaningful numbers values;
– apply those number values for computing priorities for the elements in the hierarchy;

– synthesize the results in order to establish an overall outcome.

There is a multiple criteria decision making support methodology which meets the prescription developed above. It is called the Analytic Hierarchy Process (AHP) and was developed at the Wharton School of Business by Thomas Saaty (1977). Although it is a very popular and widely implemented theory of choice, it is also controversial, thus very often validated and valuated from the perspective of its methodology. From that perspective, most recent papers, such as Grzybowski (2016); Kazibudzki (2016a); Chen et al. (2015); Pereira & Costa (2015); Linares et al. (2014); Moreno-Jiménez et al. (2014); Aguarón, Escobar & Moreno-Jiménez (2014); Lin, Kou & Ergu (2013); Brunelli, Canal & Fedrizzi (2013), unfold new research areas in this matter which should be thoroughly examined and provoke questions which should be answered, that is:

*1) Is the principal right eigenvector (REV), as the prioritization procedure (PP), necessary and sufficient for the AHP?*

*2) Is the reciprocity of the Pairwise Comparison Matrix (PCM) a reasonable condition leading to the improvement of the priority ratios estimation quality?*

*3) Are PCM consistency measures, commonly applied and embedded in the AHP, really conducive to the improvement of the priority ratios estimation quality?*

## Principles of the Analytic Hierarchy Process

### Preliminaries

The AHP seems to be the most widely used multiple criteria decision making approach in the world today. Probably, the most recent list of application oriented papers can be found in Grzybowski (2016). Actual applications in which the AHP results were accepted and used by competent decision makers can be found in: Saaty (2008); Ishizaka & Labib (2011); Ho (2008); Vaidya & Kumar (2006); Bhushan & Ria (2004); or Saaty & Vargas (2006). However, regardless of AHP popularity, the genuine methodology is also undeniably the most validated, developed and perfected contemporary methodology, see for example: Kazibudzki (2016b); Chen et al. (2015); Pereira & Costa (2015); Linares et al. (2014); Moreno-Jiménez et al. (2014); or Aguarón, Escobar & Moreno-Jiménez (2014).

The AHP allows decision makers to set priorities and make choices on the basis of their objectives, knowledge and experience in a way that is consistent with their intuitive thought process. AHP has substantial theoretical and empirical support encompassing the study of human judgmental process by cognitive psychologists. It uses the hierarchical structure of the decision problem, pairwise relative comparisons of the elements in the hierarchy, and a series of redundant judgments. This approach reduces errors and provides a measure of the consistency of judgments. The process permits accurate priorities to be derived from verbal judgments even though the words themselves may not be very precise. Thus, it is possible to use words for comparing qualitative factors and then to derive ratio scale priorities that can be combined with quantitative factors.

To make a proposed solution possible i.e. derive ratio scale priorities on the basis of verbal judgments, a scale is utilized to evaluate the preferences for each pair of items. Apparently, the most popular is Saaty's numerical scale which comprises of the integers from one (equivalent to the verbal judgment - 'equally preferred') to nine (equivalent to the verbal judgment - 'extremely preferred'), and their reciprocals. However, in conventional AHP applications it may be desirable to utilize other scales also i.e. a geometric and/or numerical scale. The former usually consists of the numbers computed in accordance with the formula $2^{n/2}$ where $n$ comprises of the integers from minus eight to eight. The latter involves arbitrary integers from one to $n$ and their reciprocals.

The first step in using AHP is to develop a hierarchy by breaking a problem down into its primary components. The basic AHP model includes the goal (a statement of the overall objective), criteria (the factors that should be considered in reaching the ultimate decision) and alternatives (the feasible alternatives that are available to achieve said ultimate goal). Although the most common and basic AHP structure consists of a goal-criteria-alternatives sequence (Fig.1). AHP can easily support more complex hierarchies. A variety of basic hierarchical structures include:
– goal, criteria, sub-criteria, scenarios, alternatives;
– goal, players, criteria, sub-criteria, alternatives;
– goal, criteria, levels of intensities, many alternatives.

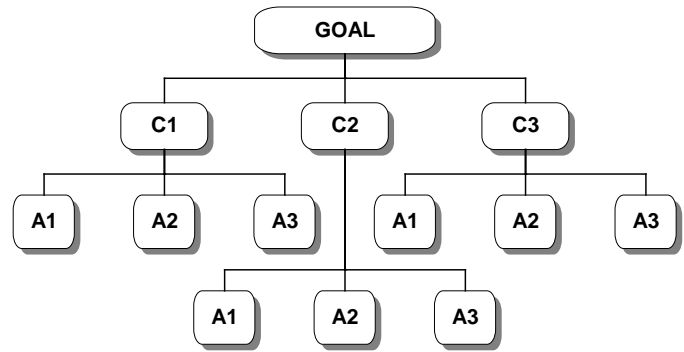

Fig. 1 - Example of a fundamental three level hierarchy encompassing three criteria and three alternatives under each criterion

## Mathematics behind the Analytic Hierarchy Process

The conventional procedure of priority ranking in AHP is grounded on the well-defined mathematical structure of consistent matrices and their associated right-eigenvector's ability to generate true or approximate weights.

The German mathematician, Oscar Perron, proved in 1907 that, if $A=(a_{ij})$, $a_{ij}>0$, where $i, j=1,\ldots, n$, then $A$ has a simple positive eigenvalue $\lambda_{max}$ called the principal eigenvalue of $A$ and $\lambda_{max}>|\lambda_k|$ for the remaining eigenvalues of $A$. Furthermore, the principal eigenvector $w=[w_1,\ldots, w_n]^T$ that is a solution of $Aw=\lambda_{max}w$ has $w_i>0$, $i=1,\ldots, n$. Thus, the conventional concept of AHP can be presented as follows:

  

$$\begin{bmatrix} w1/w1 & w1/w2 & w1/w3 & \ldots & w1/wn \\ w2/w1 & w2/w2 & w2/w3 & \ldots & w2/wn \\ w3/w1 & w3/w2 & w3/w3 & \ldots & w3/wn \\ \vdots & \vdots & \vdots & & \vdots \\ wn/w1 & wn/w2 & wn/w3 & \ldots & wn/wn \end{bmatrix} \times \begin{bmatrix} w1 \\ w2 \\ w3 \\ \vdots \\ wn \end{bmatrix} = \begin{bmatrix} nw1 \\ nw2 \\ nw3 \\ \vdots \\ nwn \end{bmatrix} \quad (1)$$

If the relative weights of a set of activities are known, they can be expressed as a Pairwise Comparison Matrix (PCM) as shown above $A(w)$. Now, knowing $A(w)$ but not $w$ (vector of priority ratios), Perron's theorem can be applied to solve this problem for $w$. The solution leads to $n$ unique values for $\lambda$, with an associated vector $w$ for each of the $n$ values.

PCMs in the AHP reflect relative weights of considered activities (criteria, scenarios, players, alternatives, etc.), so the matrix $A(w)$ has a special form. Each subsequent row of that matrix is a constant multiple of its first row. In this case a matrix $A(w)$ has only one non-zero eigenvalue, and since the sum of the eigenvalues of a positive matrix is equal to the sum of its diagonal elements, the only non-zero eigenvalue in such a case equals the size of the matrix and can be denoted as $\lambda_{max} = n$.

The norm of the vector $w$ can be written as $\|w\| = e^T w$ where: $e = [1, 1, \ldots, 1]^T$ and $w$ can be normalized by dividing it by its norm. For uniqueness, $w$ is referred to in its normalized form.

Theorem 1: A positive $n$ by $n$ matrix has the ratio form $A(w) = (w_i/w_j)$, $i, j = 1, \ldots, n$, if, and only if, it is consistent.

Theorem 2: The matrix of ratios $A(w) = (w_i/w_j)$ is consistent if and only if $n$ is its principal eigenvalue and $Aw = nw$. Further, $w > 0$ is unique to within a multiplicative constant.

Definition 1: If the elements of a matrix $A(w)$ satisfy the condition $w_{ij} = 1/w_{ji}$ for all $i, j = 1, \ldots, n$ then the matrix $A(w)$ is called *reciprocal*.

Definition 2: The matrix $A(w)$ is called *ordinal transitive* if the following conditions hold:
(A) if for any $i = 1, \ldots, n$, an element $a_{ij}$ is not less than an element $a_{ik}$ then $a_{ij} \geq a_{ik}$ for $i = 1, \ldots, n$, and
(B) if for any $i = 1, \ldots, n$, an element $a_{ji}$ is not less than an element $a_{ki}$ then $a_{ji} \geq a_{ki}$ for $i = 1, \ldots, n$.

Definition 3: If the elements of a matrix $A(w)$ satisfy the condition $w_{ik}w_{kj} = w_{ij}$ for all $i, j, k = 1, \ldots, n$, and the matrix is *reciprocal*, then it is called *consistent* or *cardinal transitive*.

Certainly, in real life situations when AHP is utilized, there is not an $A(w)$ which would reflect weights given by the vector of priority ratios. As was stated earlier, the human mind is not a reliable measurement device. Assignments such as, 'Compare – applying a given ratio scale – your feelings concerning alternative 1 versus alternative 2', do not produce accurate outcomes. Thus, $A(w)$ is not established but only its estimate $A(x)$ containing intuitive judgments, more or less close to $A(w)$ in accordance with experience, skills, specific knowledge, personal taste and even temporary mood or overall disposition. In such case, consistency property does not hold and the relation between elements of $A(x)$ and $A(w)$ can be expressed as follows:

$$x_{ij} = e_{ij} w_{ij} \tag{2}$$

where $e_{ij}$ is a perturbation factor fluctuating near unity. In the statistical approach $e_{ij}$ reflects a realization of a random variable with a given probability distribution.

It has been shown that for any matrix, small perturbations in the entries imply similar perturbations in the eigenvalues, that is why in order to estimate the true priority vector $w$, conventional AHP utilizes Perron's theorem. The solution of the matrix equation $Aw=\lambda_{max}w$, gives us $w$ as the Right Principal Eigenvector (REV) associated with $\lambda_{max}$.

In practice the REV solution is obtained by raising the matrix $A(x)$ to a sufficiently large power, then the rows of $A(x)$ are summed and the resulting vector is normalized in order to receive $w$. This concept can be also delivered in the form of the following formula:

$$w = \lim_{k \to \infty} \left( \frac{A^k \times e}{e^T \times A^k \times e} \right) \tag{3}$$

where: $e=[1, 1,\ldots, 1]^T$.

## Description of the first problem

It has been promoted that the REV prioritization procedure (PP) is necessary and sufficient to uniquely establish the ratio scale rank order inherent in inconsistent pairwise comparison judgments (Saaty & Hu 1998). However, there are alternative PPs devised to cope with this problem. Many of them are optimization based and seek a vector $w$, as a solution of the minimization problem given by the formula:

$$\min D(A(x), A(w)) \tag{4}$$

subject to some assigned constraints such as, for example, positive coefficients and normalization condition. Because the distance function $D$ measures an interval between matrices $A(x)$ and $A(w)$, different ways of its definition lead to various prioritization concepts and prioritization results. As an example, Choo et al. (2004) describes and compares eighteen estimation procedures for ranking purposes although some authors suggest there are only fifteen that are different. Furthermore, since the publication of the above article, a few additional procedures have been introduced to the literature, see for example: Grzybowski (2012).

Certainly, when the PCM is consistent, all known procedures coincide. However, in real life situations, as was discussed earlier, human judgments produce inconsistent PCMs. The inconsistency is a natural consequence of human brain dynamics described earlier and also a consequence of the questioning methodology, mistaken entering of judgment values, and scaling procedure (i.e. rounding errors). It seems crucial to emphasize here that usually even perfectly consistent PCMs, only because of rounding errors are not error-free. It can be illustrated on the basis of the following hypothetic example.

The genuine priority vector: $w=[7/20, 1/4, 1/4, 3/20]$ is considered and derived from it, $A(w)$ which can be presented as follows:

$$A(w) = \begin{bmatrix} 1 & 7/5 & 7/5 & 7/3 \\ 5/7 & 1 & 1 & 5/3 \\ 5/7 & 1 & 1 & 5/3 \\ 3/7 & 3/5 & 3/5 & 1 \end{bmatrix}$$

Now it is considered $A(x)$ produced by a hypothetic decision maker (DM), whose judgments are perfectly consistent. Even if it is assumed that the selected DM is very trustworthy and can express judgments very precisely, DM is still somehow limited by the necessity of expressing judgments on a scale (the example utilizes Saaty's scale). As such, the DM will produce the PCM ($A(x)$) which is not error-free because the entries must be in this case rounded to the closest values of Saaty's scale. Since $A(x)$ must be reciprocal (the fundamental requirement of the AHP) the PCM appears as follows:

$$A(x) = \begin{bmatrix} 1 & 1 & 1 & 2 \\ 1 & 1 & 1 & 2 \\ 1 & 1 & 1 & 2 \\ 1/2 & 1/2 & 1/2 & 1 \end{bmatrix}$$

It may be noticed that the above PCM is perfectly consistent, so this construct seems to be exemplary. However, the hypothetic DM, despite best intentions, is burdened with inescapable estimation errors. In the above situation the priority vector (PV) derived from $A(x)$ by any PP, provides the following priority ratios (PRs): $x$=[2/7, 2/7, 2/7, 1/7] which are not equal to those considered exemplary: $w$=[7/20, 1/4, 1/4, 3/20]. Obviously, the deviation between those PVs can also be expressed by their Mean Absolute Error (MAE), for instance, established by the following formula:

$$MAE(w,x) = \frac{1}{n}\sum_{i=1}^{n}|w_i - x_i| \qquad (5)$$

where $n$ is the number of elements within the particular PV. Noticeably, in the above example, MAE equals 1/28.

From that perspective, Saaty & Hu's (1998) declaration articulating that the REV *is the only valid PP for deriving the PV from a PCM, particularly when the PCM is inconsistent* seems at least questionable. However, they provide an example of a situation where variability in ranks does not occur for each individual judgment matrix, it occurs in the overall ranking of the final alternatives due to the application of different PPs and the multi-criteria process itself. They argue that only the REV possesses a sound mathematical background directly dealing with the question of inconsistency. Furthermore, as they state, only the REV captures the rank order inherent in the inconsistent data in a unique manner. It appears to be time to verify the credibility of these statements utilizing the Monte Carlo simulations.

For that purpose, apart from the REV, four different PPs have been arbitrarily selected ranked as the best within AHP methodology (Kazibudzki & Grzybowski 2013; Lin 2007; Choo & Wedley 2004) – Table 1.

Table 1 – Formulae for the prioritization procedures

| The Prioritization Procedure | Formula for the Prioritization Procedure |
|---|---|
| Logarithmic Utility Approach – LUA – | $w_{(LUA)} = \min \sum_{i=1}^{n} \ln^2 \left( \sum_{j=1}^{n} \frac{a_{ij} w_j}{n w_i} \right)$ |
| Sum of Squared Relative Differences Method – SRDM | $w_{(SRDM)} = \min \sum_{i=1}^{n} \left( \frac{1}{n w_i} \sum_{j=1}^{n} a_{ij} w_j - 1 \right)^2$ |
| Logarithmic Least Squares Method – LLSM – | $w_{(LLSM)} = \min \sum_{i=1}^{n} \sum_{j=1}^{n} \ln^2 \left( a_{ij} \frac{w_j}{w_i} \right)$ |
| Simple Normalized Column Sum – SNCS – | $w_{i(SNCS)} = \frac{1}{n} \sum_{j=1}^{n} \left( a_{ij} \bigg/ \sum_{k=1}^{n} a_{kj} \right)$ |

# The first problem study

The objective of this chapter is to verify the above statement i.e. *the REV is the only valid method for deriving the PV from a PCM, particularly when the matrix is inconsistent.*

Taking into account the exemplary study of Saaty & Hu (1998), it seems that the best way to analyze the problem is to examine whether different PPs are really inferior in the estimation of true PVs whose intent is accurate estimation. From that perspective, only computer simulations can illuminate the question, for it is possible to elaborate an algorithm which enables simulation of different kinds of errors which may occur during the process of judgment, and enables assessment which one from the selected PPs delivers better estimates (from a given perspective) of the genuine PV.

Thus, the following simulation algorithm was constructed. Assuming that the decisional problem can be presented in the form of a three level hierarchy (goal, criteria and alternatives – see Figure 1). In order to emulate the problem presented in Saaty & Hu (1998), the hypothetical hierarchy is also designed as a four criteria and four alternatives structure i.e. $n=4$ and $m=4$. In agreement with these assumptions, it is possible to elaborate and execute the simulation algorithm **SA|1|** comprising of the following steps:

***Step 1***. Randomly generate a priority vector $\mathbf{k}=[k_1,…, k_n]^T$ of assigned size [*n* x *1*] for criteria and related perfect PCM(k)=***K***(*k*)

***Step 2***. Randomly generate exactly *n* priority vectors $\mathbf{a_n}=[a_{n,1},…, a_{n,m}]$ of assigned size [*m* x *1*] for alternatives under each criterion and related perfect PCMs(*a*)=***A<sub>n</sub>***(*a*)

***Step 3***. Compute a total priority vector ***w*** of the size [*m* x *1*] applying the following procedure: $w_x=k_1 a_{1,x} + k_2 a_{2,x} +…+ k_n a_{n,x}$

***Step 4***. Randomly choose a number *e* from the assigned interval [α; β] on the basis of assigned probability distribution π

***Step 5***. Apply separately *Step 5A* and *Step 5B*:

*Step 5A* – *the case of PCM forced reciprocity implementation;*

> replace all elements $a_{ij}$ for $i<j$ of all $\boldsymbol{A_n}(a)$ with $ea_{ij}$, and all elements $k_{ij}$ for $i<j$ of $\boldsymbol{K}(k)$ with $ek_{ij}$

*Step 5B* – *the case of arbitrary PCM acceptance;*

> replace all elements $a_{ij}$ for $i \neq j$ of all $\boldsymbol{A_n}(a)$ with $ea_{ij}$, and all elements $k_{ij}$ for $i \neq j$ of $\boldsymbol{K}(k)$ with $ek_{ij}$

**Step 6**. Apply separately *Step 6A* and *Step 6B*:

*Step 6A* – *when Step 5A is performed;*

> round all values of elements $a_{ij}$ for $i<j$ of all $\boldsymbol{A_n}(a)$, and all values of elements $k_{ij}$ for $i<j$ of $\boldsymbol{K}(k)$ to the closest values from a considered scale, then replace all elements $a_{ij}$ for $i>j$ of all $\boldsymbol{A_n}(a)$ with $1/a_{ij}$, and all elements $k_{ij}$ for $i>j$ of $\boldsymbol{K}(k)$ with $1/k_{ij}$

*Step 6B* – *when Step 5B is performed;*

> round all values of elements $a_{ij}$ for $i \neq j$ of all $\boldsymbol{A_n}(a)$, and all values of elements $k_{ij}$ for $i \neq j$ of $\boldsymbol{K}(k)$ to the closest values from a considered scale

**Step 7**. On the basis of all perturbed $\boldsymbol{A_n}(a)$ denoted as $\boldsymbol{A_n}(a)^*$ and perturbed $\boldsymbol{K}(k)$ denoted as $\boldsymbol{K}(k)^*$ compute their respective priorities vectors $\boldsymbol{a_n}^*$ and $\boldsymbol{k}^*$ with application of assigned estimation procedure (EP), i.e.: REV, LUA, SRDM, LLSM, and SNCS.

**Step 8**. Compute a total priority vectors $\boldsymbol{w}^*(EP)$ of the size [$m \times 1$] applying the following procedure: $w^*_x = k^*_1 a^*_{1,x} + k^*_2 a^*_{2,x} + \ldots + k^*_n a^*_{n,x}$

**Step 9**. Calculate *Spearman rank correlation coefficients* – $SRC_{\gamma,\chi}(\boldsymbol{w}^*(EP),\boldsymbol{w})$ between all $\boldsymbol{w}^*(EP)$ and $\boldsymbol{w}$, as well designated estimation precision characteristics, i.e.: mean relative errors:

$$RE_{\gamma,\chi}(w^*(EP),w) = \frac{1}{m}\sum_{i=1}^{m}\frac{|w_i - w_i^*(EP)|}{w_i} \qquad (6)$$

along with mean relative ratios:

$$RR_{\gamma,\chi}(w^*(EP),w) = \frac{1}{m}\sum_{i=1}^{m}\frac{w_i^*(EP)}{w_i} \qquad (7)$$

**Step 10**. Repeat Steps 4 to 9, $\chi$ times, where $\chi$ denotes a size of the sample

**Step 11**. Repeat Steps 1 to 9, $\gamma$ times, where $\gamma$ denotes a number of considered AHP models

**Step 12**. Return arithmetic average values of all $SRC_{\gamma,\chi}(\boldsymbol{w}^*(EP),\boldsymbol{w})$, $RE_{\gamma,\chi}(\boldsymbol{w}^*(EP),\boldsymbol{w})$, and $RR_{\gamma,\chi}(\boldsymbol{w}^*(EP),\boldsymbol{w})$ computed during all runs in Steps: 10 and 11, i.e.:

$$MSRC(w^*(EP),w) = \frac{1}{\gamma \times \chi}\sum_{i=1}^{\gamma \times \chi} SRC_i(w^*(EP),w) \qquad (8)$$

$$MRE(w^*(EP),w) = \frac{1}{\gamma \times \chi}\sum_{i=1}^{\gamma \times \chi} RE_i(w^*(EP),w) \qquad (9)$$

$$MRR(w^*(EP),w) = \frac{1}{\gamma \times \chi}\sum_{i=1}^{\gamma \times \chi} RR_i(w^*(EP),w) \qquad (10)$$

where: $MSRC(\boldsymbol{w}^*(EP),\boldsymbol{w})$, $MRE(\boldsymbol{w}^*(EP),\boldsymbol{w})$ and $MRR(\boldsymbol{w}^*(EP),\boldsymbol{w})$ denotes: *mean Spearman rank correlation coefficient*, *average mean relative error* and *average mean relative ratio*, respectively.

In the first experiment, the probability distribution π attributed in Step 4 to the perturbation factor *e* is selected arbitrarily to be the *gamma* or *uniform* distribution. These are two of the distribution types most frequently considered in literature for various implementation purposes (Grzybowski 2016). Usually recommended are such types as: *gamma*, *log-normal*, *truncated normal*, or *uniform*. Apart from these most popular π, one

Copyright © 2016 by Paul Thaddeus KAZIBUDZKI    10/30can find applications of the Couchy, Laplace, or either *triangle* and *beta* probability distributions (see e.g. Dijkstra 2013).

The first simulation scenario also assumes that the perturbation factor *e* will be drawn from the interval $e \in [0.01; 1.99]$. Noticeably, in each case hereafter, the parameters of different probability distributions applied are set in such a way that the expected value of *e* in each particular simulation scenario EV(*e*)=1. It seems a very reasonable assumption, because although human judgments are not accurate, they undeniably aim perfect ones.

Furthermore, the number of alternatives and criteria in a single AHP model will be assigned randomly. By 'randomly' – without any other explicit specification – hereafter defined as a process operating under uniform distribution. All simulation scenarios also assume application of the rounding procedure which always operates according to the *geometric* scale described earlier in this paper.

Finally, the first scenario also takes into account the obligatory assumption in conventional AHP applications i.e. the PCM reciprocity condition. In such cases, only judgments from the upper triangle of a given PCM are taken into account and those from the lower triangle are replaced by the inverses of the former.

The outcomes i.e. mean characteristics for 30,000 cases ($\chi$=15 and $\gamma$=2000) of the first simulation scenario are presented in Table 2. It may be noticed from Table 2, that the REV can be undeniably classified as the worst PP from the perspective of PRs derived from ranks established on the basis of three different prioritization quality measures i.e. MRE, MSRC, and MRR. The best two PPs from the viewpoint of this classification are LLSM, known also as Geometric Mean Procedure (GM), and LUA. Certainly, the first scenario experiment was designed only to contrast the results presented by Saaty & Hu (1998). It is the intention to establish wider and more fundamental relationships among the presented PPs.

Table 2 – Mean performance measures of arbitrarily selected PPs for 30,000 cases

| Scenario Details | | | Procedure | MRE | Rank | MSRC | Rank | MRR | Rank | Mean Rank |
|---|---|---|---|---|---|---|---|---|---|---|
| *Geometric* Scale | *gamma* distribution | *FR*-PCM | LLSM | 0.438438 | 1 | 0.682300 | 2 | 1.21242 | 1 | 1.3(3) |
| | | | REV | 0.452614 | 5 | 0.668380 | 5 | 1.22051 | 4 | 4.6(6) |
| | | | LUA | 0.447349 | 2 | 0.673067 | 3 | 1.21792 | 2 | 2.3(3) |
| | | | SRDM | 0.448759 | 3 | 0.671380 | 4 | 1.21870 | 3 | 3.3(3) |
| | | | SNCS | 0.450734 | 4 | 0.692453 | 1 | 1.24398 | 5 | 3.3(3) |
| | *uniform* distribution | *FR*-PCM | LLSM | 0.288608 | 1 | 0.804860 | 2 | 1.12813 | 1 | 1.3(3) |
| | | | REV | 0.302346 | 4 | 0.792580 | 5 | 1.13530 | 4 | 4.3(3) |
| | | | LUA | 0.298401 | 2 | 0.795767 | 3 | 1.13350 | 2 | 2.3(3) |
| | | | SRDM | 0.299400 | 3 | 0.794820 | 4 | 1.13400 | 3 | 3.3(3) |
| | | | SNCS | 0.303463 | 5 | 0.808333 | 1 | 1.15450 | 5 | 3.6(6) |

Note: *FR*-PCM denotes *forced reciprocity* applied to PCM during simulations

The second simulation scenario was designed to encompass new assumptions not yet taken into account in the literature. First of all, taken into consideration were results obtained not only on the basis of reciprocal PCM, but also the simulation outcomes of nonreciprocal PCM. Secondly, it was decided to implement into simulations new intervals for random errors and apply their new probability distribution. As is known, many simulation analyses presented in literature assume very non symmetric intervals for a perturbation factor (considering its influence on the particular element of PCM). For example consider the interval for perturbation factor applied in the first simulation scenario i.e. $e \in [0.01;1.99]$. Under this assumption, it becomes apparent that if some entry of PCM is modified *in plus* by the perturbation factor from that particular interval, it is multiplied maximal by the number 1.99, so if the original entry is 3, the modified value will be around 6. However, if some entry of PCM is modified *in minus* by the perturbation factor from that particular interval, it may result that some entry will be multiplied by the number 0.01, so in fact the entry will be divided by 100. Thus, in the situation where the original entry is 9, the modified value will be 0.09, which can be rounded to 1/9 on the Saaty's scale. It may be noticed that this modification practically reverses the preference of DM from e.g. extremely preferred A over B, to extremely preferred B over A (applying the Saaty scale).

It is obvious that this very common assumption is imposed by another very crucial and logical assumption which states that the expected value of *e* in every particular simulation scenario should equal one i.e. $EV(e)=1$. It is quite easy to fulfill that requirement on the basis of an asymmetric interval for the perturbation factor (from the perspective of its influence on a particular element of PCM). However, it is rather a challenge to have this assumption implemented with a symmetric interval for the perturbation factor. That is why commonly applied simulation scenarios minimize the range for the perturbation factor in order to achieve at least the delusion of symmetry for $e \in [0.5;1.5]$. Nevertheless, that objective has been attained with the present research, yet to be achieved by other researchers. Presently it seems reasonable to apply symmetric intervals to simulations for the perturbation factor because they better reflect true life situations. Thus, different kinds of probability distributions (PDs) were experimented with and it was discovered that Fisher-Snedecor PD possesses the feature that can be useful in the present analysis. It occurs that for $n_1=14$ and $n_2=40$ degrees of freedom for one thousand randomly generated numbers on the basis of this PD, their mean equals 1.03617, so it is very close to unity, and these numbers fluctuate from 0.174526 to 5.57826. So, with these assumptions, we have $e \in [0.174526; 5.57826]$, which gives a very symmetric distribution for the perturbation factor, and $EV(e) \approx 1$. The results of prioritization quality for different selected PPs and assumed prioritization quality measures i.e. MSRC, MRE, and MRR obtained on the basis of described earlier simulation scenario, are presented in Table 3.

As can be noticed, the REV again is not the dominant PP from the perspective of all simulation scenarios under prescribed frameworks (it takes third place in the total classification order). Certainly, apparent differences in the PV estimation quality in relation to the selected PP are noticeable for nonreciprocal PCMs.

Then, the LUA and SRDM or LLSM dominate over the rest of the selected PPs, especially from the perspective of rank correlations which are the crucial issue from the viewpoint of rank preservation phenomena. These issues will be treated in the section entitled '*Breakthroughs and milestones of this research*'.

Table 3 – Mean performance measures of arbitrarily selected five different ranking procedures for various uniformly drawn 100,000 AHP models – 1,000 hypothetic decisional problems perturbed 100 times each (*)

| Scenario Details | | | Procedure | MRE | Rank | MSRC | Rank | MRR | Rank | Mean Rank |
|---|---|---|---|---|---|---|---|---|---|---|
| *Geometric Scale* | $n, m \in \{3, 4, ..., 7\}$ | FRPCM | LLSM | 0.123288 | 4 | 0.916281 | 1 | 1.04646 | 3 | 2.6(6) |
| | | | REV | 0.123030 | 1 | 0.915056 | 5 | 1.04546 | 1 | 2.3(3) |
| | | | LUA | 0.123044 | 3 | 0.915489 | 2 | 1.04699 | 4 | 3 |
| | | | SRDM | 0.123038 | 2 | 0.915476 | 3 | 1.04567 | 2 | 2.3(3) |
| | | | SNCS | 0.132926 | 5 | 0.915228 | 4 | 1.05865 | 5 | 4.6(6) |
| | | APCM | LLSM | 0.100511 | 1 | 0.930242 | 4 | 1.02953 | 4 | 3 |
| | | | REV | 0.101523 | 4 | 0.930164 | 5 | 1.02938 | 3 | 4 |
| | | | LUA | 0.100658 | 2 | 0.930965 | 2 | 1.02926 | 2 | **2** |
| | | | SRDM | 0.101310 | 3 | 0.930510 | 3 | 1.02925 | 1 | **2.3(3)** |
| | | | SNCS | 0.108689 | 5 | 0.931026 | 1 | 1.04315 | 5 | 3.6(6) |
| | $n, m \in \{8, 9, ..., 12\}$ | FRPCM | LLSM | 0.079748 | 4 | 0.931396 | 1 | 1.03319 | 4 | 3 |
| | | | REV | 0.079110 | 1 | 0.928266 | 5 | 1.03116 | 1 | 2.3(3) |
| | | | LUA | 0.079321 | 3 | 0.928817 | 2 | 1.03173 | 3 | 2.6(6) |
| | | | SRDM | 0.079286 | 2 | 0.928769 | 4 | 1.03166 | 2 | 2.6(6) |
| | | | SNCS | 0.086223 | 5 | 0.928799 | 3 | 1.03935 | 5 | 4.3(3) |
| | | APCM | LLSM | 0.063936 | 4 | 0.943393 | 3 | 1.02252 | 4 | 3.6(6) |
| | | | REV | 0.062735 | 3 | 0.942399 | 5 | 1.02070 | 1 | 3 |
| | | | LUA | 0.061757 | 1 | 0.944593 | 1 | 1.02109 | 3 | **1.6(6)** |
| | | | SRDM | 0.061852 | 2 | 0.944314 | 2 | 1.02105 | 2 | **2** |
| | | | SNCS | 0.068981 | 5 | 0.942764 | 4 | 1.02879 | 5 | 4.6(6) |
| *Saaty's scale* | $n, m \in \{3, 4, ..., 7\}$ | FRPCM | LLSM | 0.143650 | 4 | 0.911381 | 1 | 1.06578 | 4 | 3 |
| | | | REV | 0.142967 | 1 | 0.911151 | 4 | 1.06498 | 1 | 2 |
| | | | LUA | 0.143069 | 3 | 0.911347 | 2 | 1.06520 | 3 | 2.6(6) |
| | | | SRDM | 0.143054 | 2 | 0.911320 | 3 | 1.06517 | 2 | 2.3(3) |
| | | | SNCS | 0.155694 | 5 | 0.910735 | 5 | 1.07850 | 5 | 5 |
| | | APCM | LLSM | 0.116095 | 1 | 0.927455 | 1 | 1.04681 | 3 | **1.6(6)** |
| | | | REV | 0.116994 | 4 | 0.926955 | 4 | 1.04705 | 4 | 4 |
| | | | LUA | 0.116337 | 2 | 0.927129 | 3 | 1.04657 | 1 | **2** |
| | | | SRDM | 0.116962 | 3 | 0.926532 | 5 | 1.04658 | 2 | 3.3(3) |
| | | | SNCS | 0.127154 | 5 | 0.927397 | 2 | 1.06051 | 5 | 4 |
| | $n, m \in \{8, 9, ..., 12\}$ | FRPCM | LLSM | 0.100279 | 4 | 0.917231 | 1 | 1.04856 | 4 | 3 |
| | | | REV | 0.098084 | 1 | 0.915833 | 4 | 1.04630 | 1 | 2 |
| | | | LUA | 0.098648 | 3 | 0.916245 | 2 | 1.04695 | 3 | 2.6(6) |
| | | | SRDM | 0.098569 | 2 | 0.916193 | 3 | 1.04687 | 2 | 2.3(3) |
| | | | SNCS | 0.106674 | 5 | 0.915633 | 5 | 1.05424 | 5 | 5 |
| | | APCM | LLSM | 0.078464 | 4 | 0.938192 | 3 | 1.03563 | 4 | 3.6(6) |
| | | | REV | 0.077002 | 3 | 0.937837 | 4 | 1.03422 | 1 | 2.6(6) |
| | | | LUA | 0.076762 | 1 | 0.939669 | 1 | 1.03469 | 3 | **1.6(6)** |
| | | | SRDM | 0.076789 | 2 | 0.939415 | 2 | 1.03464 | 2 | **2** |
| | | | SNCS | 0.084307 | 5 | 0.937796 | 5 | 1.04125 | 5 | 5 |
| **Average Mean Rank** | | | LLSM | 2.958 | | REV | 2.792 | | LUA | 2.292 | SRDM 2.417 | SNCS 4.542 |
| **Order** | | | 4 | | | 3 | | | 1 | 2 | 5 |

Note: (*) AHP models drawn randomly (uniformly) for assigned set of criteria and alternatives. The scenario assumes application of both: perturbation factor drawn with F-Snedecor probability for $n_1=14$ and $n_2=40$

degrees of freedom, and rounding errors associated with a given scale (geometric or Saaty's). It assumes calculation of performance measures either for reciprocal PCMs (FRPCM) or nonreciprocal PCMs (APCM).

## Description of the second problem

In the previous two sections of this research, it was determined that the quality of PV estimation depends on the selected PP. This section will focus on the other facet of the problem i.e. how the quality of PV estimation depends on the type of PCM Consistency Measure (PCM-CM) engaged in the prioritization process. The difference between the meaning of consistency of a given PCM and the particular PCM-CM is intentionally stressed at this point. Indeed, there are several PCM-CMs provided in the literature called consistency indices (CIs), nevertheless the scientific meaning of PCM consistency is given by the definition (Definition 3).

The most popular and certainly less intuitive is the PCM-CM proposed by Saaty. He proposed his PCM-CM on the basis of his PP which involves eigenvectors and eigenvalues calculations. Thus, the indication of the fact that for the consistent PCM its $\lambda_{max}= n$, for the purpose of PCM consistency measurement, Saaty proposes his CI be determined by the following formula:

$$CI_{REV} = \frac{\lambda_{max} - n}{n-1} \quad (11)$$

where $n$ indicates the number of alternatives within the particular PCM. The significant disadvantage of this PCM-CM is the fact it can operate exclusively with reciprocal PCMs. In the case of nonreciprocal PCMs, this measure is useless (its values are meaningless) which in consequence seriously diminishes the value of the whole Saaty approach (Linares et al. 2014).

However, as mentioned earlier, there are a number of additional PCM-CMs. Some of them, as in the case of $CI_{REV}$, originate from the PPs devised for the purpose of the PV estimation process. Their distinct feature is the fact that all of them can operate equally efficiently in conditions where reciprocal and nonreciprocal PCMs are accepted. A number of them, selected on the basis of their popularity (but not only) and up-to-date nature (Kazibudzki 2016b) are presented in Table 4.

Noticeably, there are few propositions of PCM-CMs which are not connected with any PP and are devised on the basis of the PCM consistency definition (Definition 3). Koczkodaj's (1993) idea is the first to be considered. His PCM-CM is grounded on his concept of triad consistency. The notion of a triad:

Statement 1: For any three distinguished decision alternatives $A_1$, $A_2$, and $A_3$, there are three meaningful priority ratios i.e. α, β, and χ, which have their different locations in a particular $A(w)=[w_{ij}]_{n\times n}$

Definition 4: If α=$w_{ik}$, χ=$w_{kj}$, β=$w_{ij}$ for some different $i\leq n$, $j\leq n$, and $k\leq n$, then the tuple (α, β, χ) is called a *triad*.

Definition 5: If the matrix $A(w)=[w_{ij}]_{n\times n}$ is consistent, then αχ=β for all triads.

Table 4 – Formulae for the PCM-CMs related to their PPs

| Symbol of the PP | Formula for the PCM-CM |
|---|---|
| LUA | $CI_{LUA} = \dfrac{1}{n}\sqrt{\min \sum_{i=1}^{n} \ln^2\left(\sum_{j=1}^{n} \dfrac{a_{ij}w_j}{nw_i}\right)}$ |
| SRDM | $CI_{SRDM} = \sqrt{\dfrac{1}{n}\min \sum_{i=1}^{n}\left(\dfrac{1}{nw_i}\sum_{j=1}^{n} a_{ij}w_j - 1\right)^2}$ |
| LLSM | $CI_{LLSM} = \dfrac{2}{(n-1)(n-2)}\sum_{i<j}\log^2\left(\dfrac{a_{ij}w_j}{w_i}\right)$ |

In consequence, either of the equations 1–β/αχ=0 and 1–αχ/β=0 have to be true. Taking the above into consideration, Koczkodaj proposed his measure for triad inconsistency by the following formula:

$$TI(\alpha,\beta,\chi) = \min\left[\left|1-\dfrac{\beta}{\alpha\chi}\right|, \left|1-\dfrac{\alpha\chi}{\beta}\right|\right] \qquad (12)$$

Following his idea, he then proposed the following CM of any reciprocal PCM:

$$K(TI) = \max[TI(\alpha,\beta,\chi)] \qquad (13)$$

where the maximum value of TI(α,β,χ) is taken from the set of all possible triads in the upper triangle of a given PCM.

On the basis of Koczkodaj's idea of triad inconsistency, Grzybowski (2016) presented his PCM consistency measure determined by the following formula:

$$A(TI) = \dfrac{1}{N}\sum_{i=1}^{N}[TI_i(\alpha,\beta,\chi)] \qquad (14)$$

Finally, following the idea, that ln(αχ/β) = minus ln(β/αχ), Kazibudzki (2016a) redefined triad inconsistency and proposed:
– two formulae for its measurement -

$$LTI_1(\alpha,\beta,\chi) = |\ln(\alpha\chi/\beta)| \qquad (15)$$

$$LTI_2(\alpha,\beta,\chi) = \ln^2(\alpha\chi/\beta) \qquad (16)$$

– and one meaningful formula for PCM-CM -

$$A(LTI_i) = \dfrac{1}{N}\sum_{j=1}^{N}[LTI_{ij}(\alpha,\beta,\chi)] \qquad (17)$$

which can be calculated on the basis of triads from the upper triangle of the given PCM when it is reciprocal, or all triads within the given PCM when it is nonreciprocal.

## The second problem study

This section begins with the fundamental question which should encourage all researchers who deal with the problem of PR estimation quality to seek appropriate PCM consistency measurement. The question asks:
*Does a growth of the PCM consistency directly lead to the betterment of the priority vector estimation quality?*

Apparently, the answer to this question seems to be affirmative. Commonly, this is the reason why one strives to keep the consistency of the PCM at the highest possible level. However, *is it a good idea to use universally recognized PCM-CMs for this purpose?* To answer this question a preliminary analysis of the example provided and examined in the section entitled '*Description of the first problem*' can be initiated.

Thus, the genuine PV is reconsidered, $w$=[7/20, 1/4, 1/4, 3/20] and $A(w)$ derived from that PV can be presented as follows:

$$A(w)=\begin{bmatrix} 1 & 7/5 & 7/5 & 7/3 \\ 5/7 & 1 & 1 & 5/3 \\ 5/7 & 1 & 1 & 5/3 \\ 3/7 & 3/5 & 3/5 & 1 \end{bmatrix}$$

Now considering two PCMs i.e. $R(x)$ and $A(x)$ produced by a hypothetical DM, whose judgments are rounded to Saaty's scale – DM is very trustworthy and is able to express judgments very precisely. In the first scenario, entries of $A(w)$ are rounded to Saaty's scale and the entries are made reciprocal (a principal condition for a PCM in the AHP) producing:

$$R(x)=\begin{bmatrix} 1 & 1 & 1 & 2 \\ 1 & 1 & 1 & 2 \\ 1 & 1 & 1 & 2 \\ 1/2 & 1/2 & 1/2 & 1 \end{bmatrix}$$

In the second scenario only entries of $A(w)$ are rounded to Saaty's scale (nonreciprocal case) producing:

$$A(x)=\begin{bmatrix} 1 & 1 & 1 & 2 \\ 1/2 & 1 & 1 & 2 \\ 1/2 & 1 & 1 & 2 \\ 1/2 & 1/2 & 1/2 & 1 \end{bmatrix}$$

It should be noted that $R(x)$ is perfectly consistent and $A(x)$ is not. Tables 5 and 6 present selected values of the PPs related PCM-CMs (that is $CI_{REV}$, $CI_{LUA}$, and $CI_{LLSM}$) for $R(x)$ and $A(x)$ together with PVs derived from $R(x)$ and $A(x)$; Mean Absolute Errors (MAEs) [Formula (18)], among $w*$(PP) and the genuine $w$ for the case; Spearman Rank Correlation Coefficients (SRCs) among $w*$(PP) and the genuine $w$ for the case.

$$MAE(w*(PP), w) = \frac{1}{n}\sum_{i=1}^{n}|w_i - w_i*(PP)| \qquad (18)$$

Table 5 – Values of the PCM-CMs for $R(x)$ and proposed characteristics of PVs estimates (*) quality in relation to the genuine PV for the case

| PP | Estimates | Performance measures | | |
|---|---|---|---|---|
| | | CI(*PP*) | MAE | SRC |
| REV | [0.285714, 0.285714, 0.285714, 0.142857]$^T$ | 0.0 | 0.0357143 | 0.8164966 |
| LUA | [0.285714, 0.285714, 0.285714, 0.142857]$^T$ | 0.0 | 0.0357143 | 0.8164966 |
| LLSM | [0.285714, 0.285714, 0.285714, 0.142857]$^T$ | 0.0 | 0.0357143 | 0.8164966 |

(*) derived from $R(x)$ with application of a particular PP

Table 6 – Values of the PCM-CMs for $A(x)$ and proposed characteristics of PVs estimates (*) quality in relation to the genuine PV for the case

| PP | Estimates | Performance measures | | |
|---|---|---|---|---|
| | | CI(*PP*) | MAE | SRC |
| REV | [0.309401, 0.267949, 0.267949, 0.154701]$^T$ | – 0.0893164 | 0.0202995 | 1 |
| LUA | [0.306135, 0.268645, 0.268645, 0.156576]$^T$ | 0.0344483 | 0.0219326 | 1 |
| LLSM | [0.314288, 0.264284, 0.264284, 0.157144]$^T$ | 0.0400378 | 0.0178559 | 1 |

(*) derived from $A(x)$ with application of a particular PP

Surprisingly, a very interesting phenomenon can be noted on the basis of information provided in Tables 5 and 6. The nonreciprocal version of the analyzed PCM contains non-zero values for the selected PCM-CMs. In cases similar to this example, the value of Saaty's PCM-CM always becomes negative which makes it inexplicable and in consequence useless under such circumstances (as already mentioned earlier). The other two measures are positive and higher than zero which indicates that the particular PCM is not consistent. On the basis of the same indicators in the case of the reciprocal version of the analyzed PCM, its perfect consistency is apparent because all selected PCM-CMs in this case are equal to zero. However, the estimation precision measures (MAE and SRC) i.e. characteristics of the particular PV estimation quality, indicate something quite opposite. Surprisingly, apparent are smaller values of MAEs and perfect correlation of ranks between estimated and genuine PV for nonreciprocal version of the analyzed PCM. Certainly, this conclusion concerns all analyzed PPs and it is very true in the situation when the particular PCM is apparently less consistent (on the basis of selected exemplary PCM-CMs).

It has been suggested that these discoveries inevitably lead to the conclusion that the time has just come to revise the common yet erroneous approach to the PCM consistency measurement which can be described as … *the lower PCM-CM, the better PR estimation quality.*

Therefore, it becomes apparent that there are actually three significantly different consistency notions: (1) the consistency of PCM stated by Definition 3, and reflected by a value of the specific CM which in its way denotes a deviation of the analyzed PCM from its fully consistent counterpart; (2) the consistency of DM i.e. their reliability from the viewpoint of their expertise, measured by a comparison of DM judgments reflected by the particular PCM with judgments made more or less randomly; and (3) the PCM consistency stated by Definition 3 and reflected by a value of the specific CM which denotes the particular PCM applicability for PRs derivation in the way that minimizes estimation errors.

The third notion is of particular interest from the perspective of the Multiple Criteria Decision Making (MCDM) quality. The key concept of the issue was first presented by Grzybowski (2016) and enhanced by Kazibudzki (2016a). It was decided to examine the phenomenon described therein and further develop it to improve the quality of MCDM. The simulation framework for this purpose was adopted from Kazibudzki (2016a) as the only way to examine said phenomena through computer simulations. The simulation algorithm **SA|2|** thus comprises of the following phases:

*Phase 1* Generate randomly a priority vector $w=[w_1,…, w_n]^T$ of assigned size $[n \times 1]$ and related perfect PCM$(w)=K(w)$

*Phase 2* Select randomly an element $w_{xy}$ for $x<y$ of $K(w)$ and replace it with $w_{xy}e_B$ where $e_B$ is a relatively significant error, randomly drawn (*uniform* distribution) from the interval $e_B \in [2;4]$. Errors of that magnitude are basically considered as "significant", see e.g.: Grzybowski (2012), Dijkastra (2013), Lee (2007).

*Phase 3* For each other element $w_{ij}$, $i<j \leq n$ select randomly a value $e_{ij}$ for the relatively small error in accordance with the given probability distribution π (applied in equal proportions as: *gamma*, *log-normal*, *truncated normal*, and *uniform* distribution) and replace the element $w_{ij}$ with the element $w_{ij} e_{ij}$ where $e_{ij}$ is randomly drawn (*uniform* distribution) from the interval $e_{ij} \in [0,5;1,5]$

*Phase 4* Round all values of $w_{ij} e_{ij}$ for $i<j$ of $K(w)$ to the nearest value of a considered scale

*Phase 5* Replace all elements $w_{ij}$ for $i>j$ of $K(w)$ with $1/w_{ij}$

*Phase 6* After all replacements are done, return the value of the examined index as well as the estimate of the vector $w$ denoted as $w^*$(PP) with application of assigned prioritization procedure (PP). Then return the mean absolute error MAE between $w$ and $w^*$(PP). Remember values computed in this phase as one record.

*Phase 7* Repeat *Phases* from 2 to 6 $N_n$ times.

*Phase 8* Repeat *Phases* from 1 to 7 $N_m$ times.

*Phase 9* Return *all* records to one database file.

Once again, all parameters of the applied PDs – *gamma*, *log-normal*, *truncated normal*, and *uniform* – in the above simulation framework are set as previously in such a way that the expected value EV($e_{ij}$)=1.

The simulation begins from $n=4$, because simulations for $n=3$ are not interesting due to direct interrelation of considered PCM consistency measures (Bozóki & Rapcsak 2008, Dijkstra 2013). For the sake of objectivity, the simulation data is gathered in the following way: all values of selected consistency measures are split into 15 separate sets designated by the quantiles $Q$ of order $p$ from 1/15 to 14/15. The 15 intervals are defined as: the first is from 0 to the quantile of order 1/15 i.e. VRCM$_1$=[0, $Q_{1/15}$), where VRCM represents *a Value Range of the Selected PCM Consistency Measure*; the second denotes VRCM$_2$=[$Q_{1/15}$, $Q_{2/15}$), and so on… to the last one which starts from the quantile of order 14/15 and goes on to infinity i.e. VRCM$_{15}$=[ $Q_{14/15}$, ∞). The following variables are examined: Mean VRCM$_n$, average MAE within VRCM$_n$ between $w$ and $w^*$(PP), MAE quantiles of the following orders, 0.05, 0.1, 0.5, 0.9, 0.95, and relations between all of them. In the preliminary simulation program, it was decided that PP=LLSM. The application of the rounding procedure was also assumed which in this preliminary research operates according to Saaty's scale.

Lastly, the scenario takes into account the compulsory assumption in conventional AHP applications i.e. the PCM reciprocity condition. The results are based on $N_n$=20, and $N_m$=500, i.e. 10,000 cases.

In the case of a good PCM-CM, one could assume that MAE quantiles of any order should monotonically grow concurrently with the growth of the selected PCM-CM e.g. VRCM index. The same relation should occur for Mean $VRCM_n$ and average MAE for $VRCM_n$. The results of the proposed simulation framework, or any other similar simulation scenario which would contradict such a pertinent relationship would unequivocally lead to the conclusion that the examined PCM-CM does not serve its purpose.

An examination from that point of view is in order, the performance of six PCM-CMs selected from among very common or recently proposed (Fig.2): Saaty $CI_{REV}$ – (Plot A), together with Crawford & Williams $CI_{LLSM}$ – (Plot B), and Koczkodaj $K(TI)$ – (Plot C), together with Grzybowski $A(TI)$ – (Plot D), as well as Kazibudzki $A(LTI_1)$ and $CI_{LUA}$ – (Plots E-F).

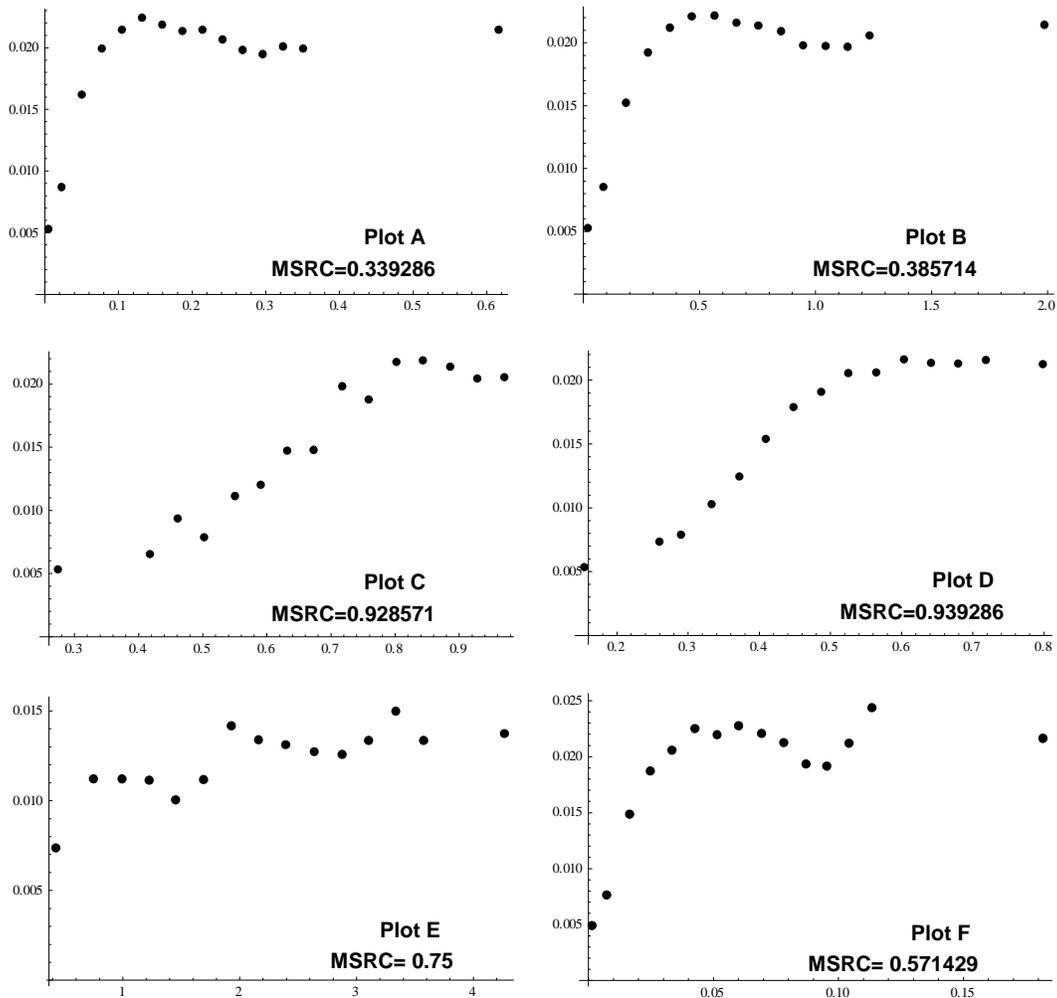

**Fig. 2 – Performance of selected PCM-CMs –** The plots present the relation between a mean value of a given PCM-CM within a given interval ($VRCM_n$) and quantiles of order 0.05 of MAEs distribution concerning estimated and genuine PV for the case. The results are generated with application of LLSM as the PP. Plots are based on 10,000 random reciprocal PCMs for $n$=4. The relation strength MSRC denotes Mean Spearman Rank Correlation Coefficient between analyzed variables.

Noticeably, when the quality of PV in MCDM process of AHP is taken into consideration, the presented relations indicate that the analyzed performance of selected PCM-CMs vary more or less from the target. Indeed, the relations indicate that most of the analyzed indices may even misinform DMs about their judgment applicability for the construct of the PV which best converge with the ideal one i.e. obtained from a fully consistent PCM. As seen similarly in the example provided earlier in this paper (Tab. 5 and 6), taking the particular index as the measure of PCM consistency, one can expect both i.e. the betterment of PRs estimation quality (increase of the estimation accuracy) together with the increase of the particular CI (decrease of PCM consistency); and inversely, the deterioration of PRs estimation quality (decrease of the estimation accuracy) together with the descent of the particular CI (improvement of PCM consistency).

Noticeably, the analyzed PCM-CMs are not selected without a reason as they are commonly applied and/or suggested as good solutions in the process of PV estimation on the basis of inconsistent PCMs (for discussion see also Grzybowski 2016). This was the motivation to search for a PCM-CM which relation to PV estimation errors, reflected by SRC, would be very close or equal to 1 (the most desirable situation).

Thus, a seminal solution is proposed in this matter. On the basis of triad inconsistency measure $LTI_2(\alpha,\beta,\chi) = \ln^2(\alpha\chi/\beta)$ introduced by Kazibudzki (2016a), the following PCM-CM is submitted:

$$CM(LTI_2) = \frac{MEAN[LTI_2(\alpha,\beta,\chi)]}{1 + MAX[LTI_2(\alpha,\beta,\chi)]} \quad (19)$$

The proposed PCM-CM is denoted as *the Triads Squared Logarithm Corrected Mean* and an examination of its performance on the basis of simulation algorithm **SA|2|** proposed earlier in this paper was carried out.

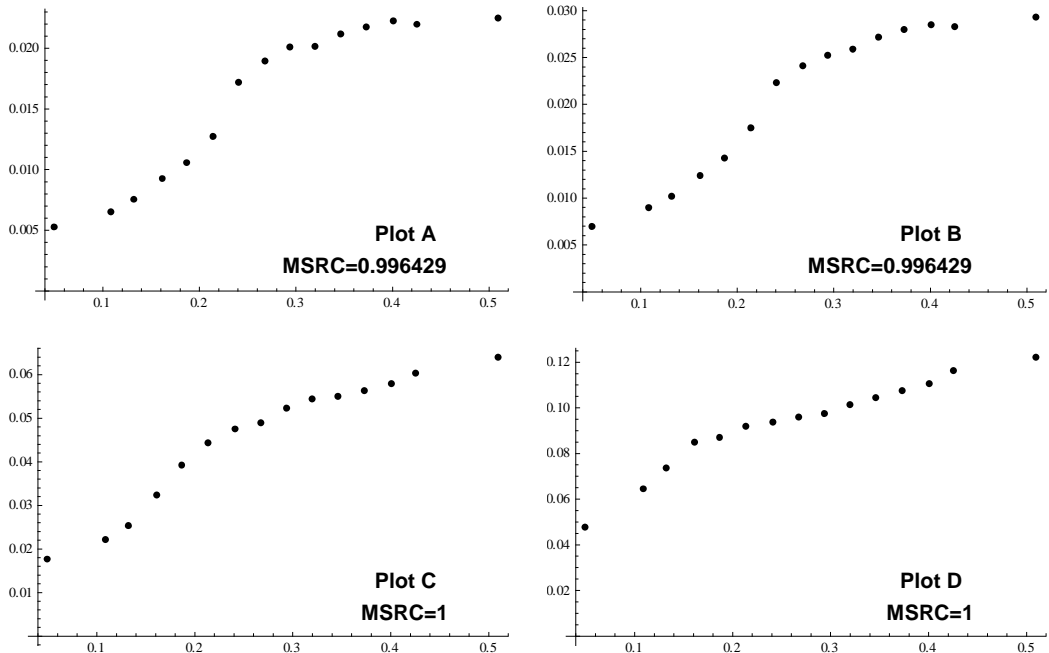

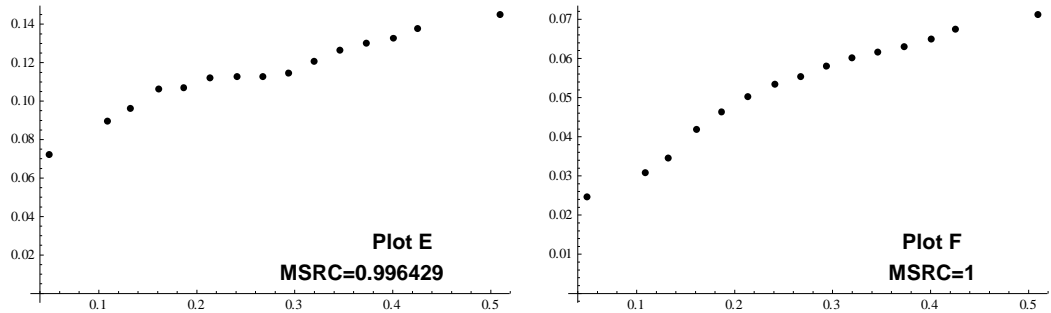

**Fig. 3 – Performance of the new PCM-CM: CM(LTI$_2$)** The plots present a relation between a mean value of CM(LTI$_2$) within a given interval (VRCM$_n$) and quantiles of order 0.05, 0.1, 0.5, 0.9, 0.95 of MAEs distribution as well their average values for estimated and genuine PVs. The results are generated with application of LLSM as the PP. Plots are based on 10,000 random reciprocal PCMs for *n*=4.

As can be noticed, the proposed CM(LTI$_2$) significantly outperforms the other PCM-CMs analyzed earlier in this paper. It is undeniably a seminal revelation that unquestionably opens a new chapter in MCDM on the basis of AHP – especially because CM(LTI$_2$) is suitable for both reciprocal and nonreciprocal PCM.

## Breakthroughs and milestones of the research

As was said in 1990 by the creator of AHP: *… there is a well-known principle in mathematics that is widely practiced, but seldom enunciated with sufficient forcefulness to impress its importance. A necessary condition that a procedure for solving a problem be a good one is that if it produces desired results, and we perturb the variables of the problem in some small sense, it gives us results that are 'close' to the original ones. (…) An extension of this philosophy in problems where order relations between the variables are important is that on small perturbations of the variables, the procedure produces close, order preserving results* (Saaty 1990, p. 18).

### The quality of PR estimation in relation to the selected PP

With said notion in mind, an effort was undertaken to verify the statement of followers of the REV, boldly spreading the idea that so long as inconsistency is accepted, the REV is the paramount theoretical basis for deriving a scale and no other concepts qualify.

It is a fact that in order to support some theory, one must verify it through many experiments to validate its reliability. On the other hand one needs only one example showing it does not work in order to abolish its credibility. Thus, numerous examples were provided indicating that the REV concedes with other devised PP to determine ranking of alternatives. However, although data obtained during simulation experiments are unequivocal, they support the above notion only generally. That is why scientific verification of their meaning is carried out on the basis of the statistical hypothesis testing theory (SHTT).

If MSRC$_{PP}$ and MSRC$_{REV}$ respectively are denoted as mean SRC of selected PP and mean SRC of the REV, their difference significance can be tested using "t" statistics defined by the following formula:

$$t = R\sqrt{\frac{n-2}{1-R^2}} \qquad (20)$$

where $R$ is the difference between particular MSRCs.

This statistic has a distribution of *t–student* with $n$ minus 2 degrees of freedom *df*, where $n$ equals the size of the sample. The following hypothesis was tested:

$H_0$: $MSRC_{PP} - MSRC_{REV} = 0$

versus

$H_1$: $MSRC_{PP} - MSRC_{REV} > 0$

In order to conform to the example presented by Saaty & Hu (1998), the data gathered in Table 2 was considered. The simulation framework of that case is *df*=29,998. Thus, for assumed levels of significance α=0.01, α=0.02 or α=0.03, the critical values of *t–student* statistics equal consecutively $t_{0.01} = 2.326472$, $t_{0.02} = 2.053838$, or $t_{0.03} = 1.880865$.

In the situation when a level of tested *t–student* statistics is higher than its critical value for the assumed level of significance, the hypothesis $H_0$, must be rejected in favor of alternative hypothesis $H_1$. In the opposite situation, there are no foundations to reject $H_0$. The selected statistics and their values for the problem evaluation are presented in Table 7.

Clearly, the results of the simulation scenario, designed in accordance with the framework presented in Saaty & Hu (1998), indicate two PPs which on the basis of SHTT always perform better than the REV, regardless of the preselected PD. It should be emphasized that the performance of selected PPs is examined here from the perspective of rank preservation phenomena which is reflected in our research by the MSRC between genuine and perturbed PV. It should be evident that the above conclusions, unlike any other before, are the effect of sound statistical reasoning (rigorous significance level) based on the seminal approach toward AHP methodology evaluation grounded on precisely planned and performed simulation study.

Table 7 – MSRC values and principal statistics for the performance test of the REV versus other selected PPs

| Scenario details | | | Procedure | MSRC | $R$ | $R^2$ | *t–value* | *α–level** |
|---|---|---|---|---|---|---|---|---|
| *Geometric* Scale | *gamma* distribution | *FR*-PCM | LLSM | 0.682300 | 0.01392 | 0.00019 | 2.411167969 | 0.01 |
| | | | REV | 0.668380 | >< | >< | >< | >< |
| | | | LUA | 0.673067 | 0.00469 | 0.00002 | 0.811794069 | – |
| | | | SRDM | 0.671380 | 0.00300 | 0.00001 | 0.519600260 | – |
| | | | SNCS | 0.692453 | 0.02407 | 0.00058 | 4.170635557 | 0.01 |
| | *uniform* distribution | *FR*-PCM | LLSM | 0.804860 | 0.01228 | 0.00015 | 2.127047876 | 0.02 |
| | | | REV | 0.792580 | >< | >< | >< | >< |
| | | | LUA | 0.795767 | 0.00319 | 0.00001 | 0.551988995 | – |
| | | | SRDM | 0.794820 | 0.00224 | 0.00001 | 0.387967421 | – |
| | | | SNCS | 0.808333 | 0.01575 | 0.00025 | 2.728747286 | 0.01 |

Note: (*) the closest significance level providing the ground to reject a tested hypothesis

In order to develop the concept further it was decided to expand the simulation program. The results of this endeavor are presented in Table 3. They should be considered as surprising, especially when one realizes that the PP embedded in the AHP merely takes third place in the overall performance ranking. The ranking takes into account not only MSRC, but MRE and MRR also, the latter never taken into consideration in previous simulation research. The MRR will now be examined to expand its concept and highlight its novelty.

Let's consider a vector $k$ of values to be estimated, $k$=[3, 3, 3, 3], and three of its estimates, $k_1$=[2, 4, 2, 4], $k_2$=[2, 2, 2, 2], $k_3$=[4, 4, 4, 4]. It may be noted that the MREs of all the estimates (given by formula (6)) are the same and equal 1/3. However, MRRs of the estimates (given by formula (7)) are not the same and equal respectively, $MRR_1(k,k_1)$=1, $MRR_2(k,k_2)$=2/3, $MRR_3(k,k_3)$=4/3. Obviously, the goal of estimation is both i.e. to minimize MREs and maintain the MRRs close to unity. This prerequisite is of great importance when one deals with PVs i.e. vectors normalized to unity, as in the case of AHP. Certainly, one can encounter the following three estimation scenarios.

Scenario 1    Consider a vector $w$ of genuine PRs trying to estimate $w$=[0.25, 0.25, 0.25, 0.25], and its estimate $w_1$=[0.01, 0.49, 0.05, 0.45]. This scenario gives a rather high MRE of 0.88, which indicates the mean 88% volatility of estimated PRs in relation to their primary values, and MRR=1.

Scenarios 2–3
               Consider a vector $p$ of genuine PRs trying to estimate $p$=[0.1, 0.2, 0.3, 0.4], and its two estimates $p_1$=[0.15, 0.3, 0.25, 0.3], and $p_2$=[0.05, 0.1, 0.35, 0.5]. This situation entails a moderate MRE of 0.35425 for both estimates, and two MRRs i.e. $MRR_1(p,p_1)$=1.145, and $RR_2(p,p_2)$=0.85425, for the second and third scenario respectively.

Obviously, during the PRs estimation process, it is desirable to avoid situations exemplified by the first and second scenario. Noticeably, they both have something in common. Apart from estimation discrepancies they lead to rank reversal of the initial priorities (emphasis added).

Turning back to Table 3, having in mind the imposed simulation scenario, F-Snedecor PD mean value of a perturbation factor EV($e$)=1.03617, we can conclude as follows:
1) the applied measures (MRE, MSRC, MRR) reflecting the quality of PR estimation process within the simulation framework are always better for nonreciprocal PCMs in relation to their reciprocal equivalents;
2) the applied measures of the quality of PR estimation within the simulation framework indicate better estimation results for a relatively higher number of alternatives;
3) both MRE and MRR values indicate that the quality of PR estimation within the simulation framework is better when geometric scale is implemented instead of Saaty's scale for preferences expression of DMs (MRR is then more often less than 1.03617 which indicates less risk of rank reversal);
4) and last but not least, the REV procedure IS NOT a dominating procedure during PR estimation in the simulated framework of the AHP.

### The quality of PR estimation in relation to the CM of the PCM

Thus far the alterability of prioritization quality in consequence of the application of preselected PP, preference scale and reciprocal or nonreciprocal PCM in the AHP has been dealt with. This chapter endeavors to focus and conclude the findings concerning the alterability of prioritization quality in relation to the applied method of the PCM consistency measurement.

Figure 2 demonstrates the basic relation between the distribution of estimation MAEs and values of selected PCM-CMs when LLSM is applied as the PP. The objective was to realize that those measures are not a good indicator of the quality of PR estimation, although the quality of PR estimation should be the core of PCM consistency measurement. Thus, a seminal solution for this problem was introduced i.e. the novel PCM-CM - CM($LTI_2$) and depicted its performance in relation to the quality of PR estimation (Fig. 3). As noted, its performance is much better than the PCM-CMs presented earlier (Fig. 2), independently of the MAEs distribution characteristics applied. Below (Tables 8–9), detailed characteristic data is presented for CM($LTI_2$) for LLSM and LUA as the PPs, and Saaty's scale as the preferred applied scale.

Table 8 – Performance of the CM($LTI_2$) index Statistical characteristics of the MAEs distribution in relation to various VRCM$_i$ for $i=1,\ldots,15$ of CM($LTI_2$) values. The results were generated for $n=4$ on the basis of **SA|2|** as the simulation algorithm and are based on 10,000 perturbed random reciprocal PCMs. The scenario assumed LLSM as the PP.

| $i$ | VRCM$_i$ for CM($LTI_2$) | Mean CM($LTI_2$) in VRCM$_i$ | $p$–quantiles of MAEs among $w$ and $w$*(LLSM) | | | | | Average MAEs among $w$ and $w$*(LLSM) |
|---|---|---|---|---|---|---|---|---|
| | | | $p$=0.05 | $p$=0.1 | $p$=0.5 | $p$=0.9 | $p$=0.95 | |
| 1 | [0.00, 0.0934) | 0.049049 | 0.0052533 | 0.0070072 | 0.0177090 | 0.0478280 | 0.0722714 | 0.0246076 |
| 2 | [0.0934, 0.12) | 0.108368 | 0.0065427 | 0.0089647 | 0.0221881 | 0.0646356 | 0.0896474 | 0.0307832 |
| 3 | [0.120, 0.147) | 0.131977 | 0.0075604 | 0.0101952 | 0.0254009 | 0.0735568 | 0.0962640 | 0.0346824 |
| 4 | [0.147, 0.173) | 0.161289 | 0.0092812 | 0.0124076 | 0.0323969 | 0.0848569 | 0.1062240 | 0.0418424 |
| 5 | [0.173, 0.200) | 0.186567 | 0.0106050 | 0.0142825 | 0.0392350 | 0.0872419 | 0.1070400 | 0.0463689 |
| 6 | [0.200, 0.227) | 0.213651 | 0.0127312 | 0.0174795 | 0.0443425 | 0.0921171 | 0.1121160 | 0.0503101 |
| 7 | [0.227, 0.253) | 0.240868 | 0.0171655 | 0.0223103 | 0.0474780 | 0.0939184 | 0.1129280 | 0.0534051 |
| 8 | [0.253, 0.280) | 0.267645 | 0.0189530 | 0.0241065 | 0.0489027 | 0.0959089 | 0.1126270 | 0.0554222 |
| 9 | [0.280, 0.307) | 0.293803 | 0.0200809 | 0.0252443 | 0.0523480 | 0.0975035 | 0.1147230 | 0.0580895 |
| 10 | [0.307, 0.333) | 0.319702 | 0.0201740 | 0.0259357 | 0.0544712 | 0.1014610 | 0.1208420 | 0.0601639 |
| 11 | [0.333, 0.360) | 0.345876 | 0.0211796 | 0.0271488 | 0.0550490 | 0.1043660 | 0.1267140 | 0.0615576 |
| 12 | [0.360, 0.387) | 0.372744 | 0.0217402 | 0.0279791 | 0.0563253 | 0.1076280 | 0.1302670 | 0.0630527 |
| 13 | [0.387, 0.413) | 0.400500 | 0.0222736 | 0.0284786 | 0.0579657 | 0.1105020 | 0.1326590 | 0.0649738 |
| 14 | [0.413, 0.440) | 0.425325 | 0.0219914 | 0.0282637 | 0.0603546 | 0.1163910 | 0.1378310 | 0.0674297 |
| 15 | [0.440, ∞) | 0.509413 | 0.0224786 | 0.0293611 | 0.0639097 | 0.1220180 | 0.1448450 | 0.0711265 |

Noted, all statistical characteristics of the MAEs distribution in relation to various VRCM$_i$ for $i=1,\ldots,15$ of CM($LTI_2$) values monotonically grow in both cases. This phenomenon ascertains that the proposed measure of the quality of PR estimation in relation to PCM-CM outperforms other commonly known or recently introduced means of PCM consistency control which were examined during this research. The paramount position of the CM($LTI_2$) is additionally strengthened by the fact that its performance improves significantly for higher numbers of alternatives without regard to which PP is employed.

It should be noted that all characteristics presented herein are of great importance in MCDM, because one has to consider the potential of rejecting a "good" PCM, and vice versa i.e. the possibility of acceptance a "bad" PCM, as in the classic SHTT. However, for first time in the course of the AHP development history, the possibility of selecting the level of trustworthiness and basing decisions on statistical facts has been demonstrated. For instance, considering some hypothetic PCM for $n=4$, with its $CM(LTI_2) \approx 0.319702$ (Tab. 8), one can expect with 95% certainty that MAE should not exceed the value of 0.1208420.

Table 9 – Performance of the $CM(LTI_2)$ index Statistical characteristics of the MAEs distribution in relation to various $VRCM_i$ for $i=1,\ldots,15$ of $CM(LTI_2)$ values. The results were generated for $n=4$ on the basis of **SA|2|** as the simulation algorithm and are based on 10,000 perturbed random reciprocal PCMs. The scenario assumed LUA as the PP.

| $i$ | $VRCM_i$ for $CM(LTI_2)$ | Mean $CM(LTI_2)$ in $VRCM_i$ | $p$–quantiles of MAEs among $w$ and $w^*$(LUA) | | | | | Average MAEs among $w$ and $w^*$(LUA) |
|---|---|---|---|---|---|---|---|---|
| | | | $p=0.05$ | $p=0.1$ | $p=0.5$ | $p=0.9$ | $p=0.95$ | |
| 1 | [0.00, 0.0921) | 0.0483805 | 0.0051862 | 0.0070132 | 0.0176693 | 0.0485092 | 0.0721248 | 0.0246818 |
| 2 | [0.0921, 0.119) | 0.107336 | 0.0065804 | 0.0087362 | 0.0223436 | 0.0668610 | 0.0901757 | 0.0310714 |
| 3 | [0.119, 0.145) | 0.130827 | 0.00728515 | 0.0096983 | 0.0248230 | 0.0756282 | 0.0986153 | 0.0345387 |
| 4 | [0.145, 0.172) | 0.159831 | 0.0097014 | 0.0126492 | 0.0318836 | 0.0839675 | 0.1048160 | 0.0417584 |
| 5 | [0.172, 0.199) | 0.185763 | 0.0108996 | 0.0147705 | 0.0390121 | 0.0867685 | 0.1087370 | 0.0464742 |
| 6 | [0.199, 0.226) | 0.212789 | 0.0127518 | 0.0171452 | 0.0444749 | 0.0906489 | 0.1110220 | 0.0502253 |
| 7 | [0.226, 0.252) | 0.239711 | 0.0168641 | 0.0221950 | 0.0483727 | 0.0943307 | 0.1121290 | 0.0538373 |
| 8 | [0.252, 0.279) | 0.266664 | 0.0191223 | 0.0243966 | 0.0499312 | 0.0963741 | 0.1128810 | 0.0561933 |
| 9 | [0.279, 0.306) | 0.292923 | 0.0210745 | 0.0265733 | 0.0536876 | 0.0971178 | 0.1136750 | 0.0590709 |
| 10 | [0.306, 0.332) | 0.318738 | 0.0222280 | 0.0280330 | 0.0570706 | 0.1018000 | 0.1224680 | 0.0622836 |
| 11 | [0.332, 0.359) | 0.344798 | 0.0229873 | 0.0290093 | 0.0582741 | 0.1054570 | 0.1267530 | 0.0640174 |
| 12 | [0.359, 0.386) | 0.371865 | 0.0237677 | 0.0299580 | 0.0592460 | 0.1080910 | 0.1309080 | 0.0652984 |
| 13 | [0.386, 0.412) | 0.399489 | 0.0243569 | 0.0309199 | 0.0612529 | 0.1118710 | 0.1350460 | 0.0678424 |
| 14 | [0.412, 0.439) | 0.424271 | 0.0245079 | 0.0311770 | 0.0630208 | 0.1197740 | 0.1443030 | 0.0707793 |
| 15 | [0.439, ∞) | 0.507848 | 0.0240355 | 0.0310822 | 0.0660800 | 0.1264270 | 0.1500270 | 0.0737878 |

At the same time, one can expect with 95% certainty that it will be higher than 0.0201740. Whether one decide to accept such a PCM or reject it, obviously depends on the quality requirements of PR estimation and the attitude regarding these errors. Indeed, the outcome of the research finally creates the potential for true consistency control in an unprecedented way i.e. directly related to the PR estimation quality.

Consider the following PV as $w=[0.345, 0.335, 0.32]$ of DM preferences for alternatives, $A_1$, $A_2$, $A_3$, respectively. Taking into consideration earlier assumed level of $CM(LTI_2) \approx 0.319702$, the order of alternatives ranks i.e. $A_1=1$, $A_2=2$, $A_3=3$, can be very deceptive, and is rather meaningless. Indeed, in such a situation one can expect with 95% certainty that MAE>0.0201740 which makes one aware that the true rank order of examined preferences may look otherwise, due to estimation errors related to DM inconsistency e.g. $w=[(0.345–0.04), (0.335+0.01), (0.32+0.03)]=[0.305, 0.345, 0.35]$, which designates a different order for alternatives ranks, $A_1=3$, $A_2=2$, $A_3=1$.

On the other hand, consider PV as $w=[0.6, 0.35, 0.05]$ of DM preferences for alternatives: $A_1$, $A_2$, $A_3$, consecutively, as previously. Again, assuming $CM(LTI_2) \approx 0.319702$, it can be anticipated with 95% certainty that MAE<0.1208420 which insures confidence in the order of alternatives ranks.

In order to conserve the length of the paper, but at the same time enable similar analyses concerning different numbers of alternatives the exemplary generalized (results are averaged for geometric scale and Saaty's scale applied fifty-fifty) characteristics of $CM(LTI_2)$ performance for $n>4$, and for selected PP in appendices to this article are provided (Tables: A1–A2).

Concluding, this simulation framework a performance of different PCM-CMs in relation to implementation of the most popular PPs, preference scales, and number of alternatives were compared. The research findings can be stated as follows:
1) it is possible to significantly improve the quality of PR estimation when CM($LTI_2$) is applied as the PCM-CM;
2) LLSM and LUA as the PP, differ insignificantly from the perspective of CM($LTI_2$) performance, the same concerns other examined PP;
3) when the number of alternatives grows, the performance of examined PCM-CMs improves.

## Conclusions and further research areas

The objective of the article was to generate answers to the following questions:
*Is the REV as the PP necessary and sufficient for the AHP? Is the reciprocity of PCMs a reasonable condition leading to the betterment of the PRs estimation quality? Are PCM-CMs, commonly applied and embedded in the AHP, really conducive to the improvement of the PRs estimation quality?*

The thorough and seminal investigation which significantly upgrades the AHP methodology provides the following answers to these questions:
1) the REV as the PP is not necessary and sufficient for the AHP. Moreover, the research reveals two PP which outperform the REV;
2) the reciprocity of PCM in the AHP is the artificial condition and directly leads to deterioration of the PR estimation quality.
3) the commonly applied PCM-CMs embedded in the AHP, mislead and in consequence often directly lead to deterioration of the PR estimation quality.

Proposed: resign from known PCM-CMs embedded in the AHP in favor of CM($LTI_2$) that can operate both types of PCM i.e. reciprocal and nonreciprocal, withhold the PCM reciprocity requirement from the AHP and consider the replacement of the REV as the PP within the AHP in favor of LUA or LLSM.

Certainly, there is a need for further research in the field. Firstly, one should examine the performance of CM($LTI_2$) when nonreciprocal PCM are applied. Secondly, one may study its performance from the perspective of relative estimation errors, and last but not least, one could evaluate its performance from the perspective of the entire hierarchy as opposed to a single PCM.

To recapitulate; in conjunction with other contemporary and seminal research papers e.g. Grzybowski (2016); Kazibudzki (2016a, 2016b); García-Melón et al. (2016); Chen et al. (2015); Pereira & Costa (2015); Linares et al. (2014); Moreno-Jiménez et al. (2014); Aguarón, Escobar & Moreno-Jiménez (2014); Lin, Kou & Ergu (2013); Brunelli, Canal & Fedrizzi (2013), the results of this scientific research enriches the state of knowledge about the true value of the AHP which is widely recognized as an applicable MCDM support system. Hopefully, the results of this freshly finished authentic examination will improve the quality of human's prospective choices.

References

Aguarón, J., Escobar, M.T., Moreno-Jiménez, J.M. (2014). The precise consistency consensus matrix in a local AHP-group decision making context, *Ann. Oper. Res.*, 1–15; http://dx.doi.org/10.1007/s10479-014-1576-8.

Aguaron, J., Moreno-Jimenez, J.M. (2003). The geometric consistency index: Approximated thresholds. *Euro. J. Oper. Res.* 147, 137–145; http://dx.doi.org/10.1016/S0377-2217(02)00255-2.

Bhushan, N., Ria, K. (2004). *Strategic Decision Making: Applying the Analytic Hierarchy Process*. Springer-Verlag London Limited, London.

Blumenthal, A.L. (1977). *The Process of Cognition*. Prentice Hall, Englewood Cliffs, New York.

Brunelli, M., Canal, L., Fedrizzi, M. (2013). Inconsistency indices for pairwise comparison matrices: a numerical study, *Ann. Oper. Res.*, 211(1), 493–509; http://dx.doi.org/10.1007/s10479-013-1329-0.

Caballero, R., Romero, C., Ruiz, F. (2016). Multiple criteria decision making and economics: an introduction, *Ann. Oper. Res.,* 245(1), 1–5; http://dx.doi.org/10.1007/s10479-016-2287-0.

Chen, K., Kou, G., Tarn, J.M., Song, Y. (2015). Bridging the gap between missing and inconsistent values in eliciting preference from pairwise comparison matrices, *Ann. Oper. Res.*, 1–21; http://dx.doi.org/10.1007/s10479-015-1997-z.

Choo, E.U., Wedley, W.C. (2004). A common framework for deriving preference values from pairwise comparison matrices. *Comp. Oper. Res.* 31, 893–908; http://dx.doi.org/10.1016/S0305-0548(03)00042-X.

Dijkstra, T.K. (2013). On the extraction of weights from pairwise comparison matrices, *Cent. Euro. J. Oper. Res.*, 21(1), 103–123; http://dx.doi.org/10.1007/s10100-011-0212-9.

García-Melón, M., Pérez-Gladish, B., Gómez-Navarro, T., Mendez-Rodriguez, P. (2016). Assessing mutual funds' corporate social responsibility: a multi-stakeholder-AHP based methodology, *Ann. Oper. Res.,* 244(2), 475–503; http://dx.doi.org/10.1007/s10479-016-2132-5.

Grzybowski, A.Z. (2016). New results on inconsistency indices and their relationship with the quality of priority vector estimation, *Expert Syst. Appl.*, 43, 197–212; http://dx.doi.org/10.1016/j.eswa.2015.08.049.

Grzybowski, A.Z. (2012). Note on a new optimization based approach for estimating priority weights and related consistency index. *Expert Syst. Appl.,* 39, 11699–11708; http://dx.doi.org/10.1016/j.eswa.2012.04.051.

Ho, W. (2008). Integrated analytic hierarchy process and its applications – A literature review, *Euro. J. Oper. Res.*, 186, 211–228; http://dx.doi.org/10.1016/j.ejor.2007.01.004.

Ishizaka, A., Labib, A. (2011). Review of the main developments in the analytic hierarchy process, *Expert Syst. Appl.,* 11(38), 14336–14345; http://dx.doi.org/10.1016/j.eswa.2011.04.143.

Kazibudzki, P. (2016a). Redefinition of triad's inconsistency and its impact on the consistency measurement of pairwise comparison matrix, *Journal of Applied Mathematics and Computational Mechanics*, 15(1), 71–78; http://dx.doi.org/10.17512/jamcm.2016.1.07.

Kazibudzki, P. (2016b). An examination of performance relations among selected consistency measures for simulated pairwise judgments, *Ann. Oper. Res.*, 244(2), 525–544; http://dx.doi.org/10.1007/s10479-016-2131-6).

Kazibudzki, P.T., Grzybowski, A.Z. (2013). On some advancements within certain multicriteria decision making support methodology, *American Journal of Business and Management,* 2(2), 143–154; http://dx.doi.org/10.11634/216796061302287.

Koczkodaj, W.W. (1993). A new definition of consistency of pairwise comparisons, *Mathematical and Computer Modeling,* 18(7), 79–84; http://dx.doi.org/10.1016/0895-7177(93)90059-8

Lin, C., Kou, G., Ergu, D. (2013) An improved statistical approach for consistency test in AHP, *Ann. Oper. Res.*, 211(1), 289–299; http://dx.doi.org/10.1007/s10479-013-1413-5.

Lin, C. (2007). A revised framework for deriving preference values from pairwise comparison matrices. *Euro. J. Oper. Res.*, 176, 1145–1150; http://dx.doi.org/10.1016/j.ejor.2005.09.022.

Linares, P., Lumbreras, S., Santamaría, A., Veiga, A. (2014). How relevant is the lack of reciprocity in pairwise comparisons? An experiment with AHP, *Ann. Oper. Res.*, 1–18; http://dx.doi.org/10.1007/s10479-014-1767-3.

Martin, J. (1973). *Design of Man-Computer Dialogues*. Prentice Hall, Englewood Cliffs, New York.

Miller, G.A. (1956). The magical number seven, plus or minus two: some limits on our capacity for information processing. *Psychol. Review*, 63, 81–97; http://dx.doi.org/10.1037/0033-295X.101.2.343.

Moreno-Jiménez, J.M., Salvador, M., Gargallo, P., Altuzarra, A. (2014). Systemic decision making in AHP: a Bayesian approach, *Ann. Oper. Res.*, 1–24; http://dx.doi.org/10.1007/s10479-014-1637-z.

Pereira, V., Costa, H.G. (2015). Nonlinear programming applied to the reduction of inconsistency in the AHP method, *Ann. Oper. Res.*, 229(1), 635–655; http://dx.doi.org/10.1007/s10479-014-1750-z.

Saaty, T.L. (2008). Decision making with the analytic hierarchy process, *Int. J. Services Sciences*, 1(1), 83–98; http://dx.doi.org/10.1504/IJSSci.2008.01759.

Saaty, T.L., Hu, G. (1998). Ranking by Eigenvector versus other methods in the Analytic Hierarchy Process, *Appl. Math. Lett.*, 11(4), 121–125; http://dx.doi.org/10.1016/S0893-9659(98)00068-8.

Saaty, T.L. (1977). A scaling method for priorities in hierarchical structures, *Journal of Mathematical Psychology*, 15, 234–81; http://dx.doi.org/10.1016/0022-2496 (77)90033-5

Saaty, T.L., Vargas, L.G. (2006). *Decision Making with the Analytic Network Process: Economic, Political, Social and Technological Applications with Benefits, Opportunities, Cost and Risks*. Springer, New York.

Saaty, T.L. (2000). *The Brain: Unraveling the Mystery of How it Works.* RWS Publications, Pittsburgh, PA.

Saaty, T.L. (1993). *The Hierarchon.* RWS Publication, Pittsburgh, PA.

Vaidya, O.S., Kumar, S. (2006). Analytic hierarchy process: An overview of applications, *Euro. J. Oper. Res.*, 169, 1–29; http://dx.doi.org/10.1016/j.ejor.2004.04.028.

Appendices

**Table A1 – Performance of CM($LTI_2$) index under the action of LLSM as the PP**. Statistical characteristics of the MAEs distribution in relation to various levels of CM($LTI_2$) within a given VRCM$_i$ for $i$=1,…,15. The results are based on 10,000 perturbed random reciprocal PCMs (geometric and Saaty's scales applied fifty-fifty), and were generated on the basis of **SA|2|** as the simulation algorithm. The table contains results for $n \in \{5, 6, 7, 8, 9\}$, presented consecutively.

| $i$ | VRCM$_i$ for CM($LTI_2$) | Mean CM($LTI_2$) in VRCM$_i$ | $p$–quantiles of MAEs among **w** and **w**\*(LLSM) | | | | | Average MAEs among **w** and **w**\*(LLSM) |
|---|---|---|---|---|---|---|---|---|
| | | | $p$=0.05 | $p$=0.1 | $p$=0.5 | $p$=0.9 | $p$=0.95 | |
| 1 | [0, 0.0899) | 0.057912 | 0.0039186 | 0.0049954 | 0.0109799 | 0.0221753 | 0.0274887 | 0.0127898 |
| 2 | [0.0899, 0.107) | 0.099124 | 0.0056158 | 0.0073876 | 0.0157136 | 0.0324243 | 0.0398139 | 0.0183201 |
| 3 | [0.107, 0.124) | 0.116088 | 0.0063140 | 0.0079525 | 0.0184299 | 0.0389673 | 0.0490687 | 0.0214159 |
| 4 | [0.124, 0.142) | 0.133907 | 0.0075132 | 0.0102233 | 0.0230429 | 0.0443459 | 0.0539668 | 0.0258898 |
| 5 | [0.142, 0.159) | 0.151127 | 0.0099921 | 0.0129535 | 0.0261046 | 0.0486044 | 0.0581258 | 0.0290851 |
| 6 | [0.159, 0.176) | 0.167911 | 0.0113191 | 0.0142543 | 0.0289546 | 0.0558904 | 0.0682777 | 0.0328936 |
| 7 | [0.176, 0.193) | 0.184671 | 0.0125612 | 0.0158052 | 0.0320054 | 0.0594491 | 0.0730399 | 0.0357402 |
| 8 | [0.193, 0.211) | 0.201896 | 0.0136853 | 0.0171375 | 0.0339101 | 0.0640703 | 0.0789391 | 0.0380755 |
| 9 | [0.211, 0.228) | 0.219329 | 0.0142803 | 0.0178080 | 0.0361548 | 0.0711273 | 0.0839402 | 0.0408705 |
| 10 | [0.228, 0.245) | 0.236371 | 0.0150518 | 0.0185369 | 0.0380656 | 0.0762136 | 0.0919801 | 0.0435024 |
| 11 | [0.245, 0.262) | 0.253302 | 0.0161087 | 0.0208189 | 0.0405464 | 0.0789105 | 0.0929572 | 0.0462684 |
| 12 | [0.262, 0.280) | 0.270523 | 0.0160427 | 0.0205586 | 0.0431223 | 0.0821647 | 0.0965329 | 0.0482168 |
| 13 | [0.280, 0.297) | 0.288211 | 0.0165698 | 0.0209757 | 0.0457022 | 0.0865715 | 0.100490 | 0.0504072 |
| 14 | [0.297, 0.314) | 0.305099 | 0.0177870 | 0.0226112 | 0.0455671 | 0.0859316 | 0.100544 | 0.0507868 |
| 15 | [0.314, ∞) | 0.357080 | 0.0186614 | 0.0241816 | 0.0493007 | 0.0932224 | 0.107664 | 0.0547348 |
| $I$ | VRCM$_i$ for CM($LTI_2$) | Mean CM($LTI_2$) in VRCM$_i$ | $p$–quantiles of MAEs among **w** and **w**\*(LLSM) | | | | | Average MAEs among **w** and **w**\*(LLSM) |
| | | | $p$=0.05 | $p$=0.1 | $p$=0.5 | $p$=0.9 | $p$=0.95 | |
| 1 | [0, 0.0901) | 0.0618775 | 0.0036042 | 0.0044511 | 0.0090509 | 0.0175099 | 0.0212066 | 0.0102634 |
| 2 | [0.0901, 0.102) | 0.096694 | 0.00584585 | 0.0072472 | 0.0147877 | 0.0255157 | 0.0297767 | 0.0158427 |
| 3 | [0.102, 0.115) | 0.109186 | 0.0071783 | 0.0088408 | 0.0167774 | 0.0304119 | 0.0360603 | 0.0186253 |
| 4 | [0.115, 0.127) | 0.121228 | 0.00831565 | 0.0102091 | 0.0192100 | 0.0349865 | 0.0420209 | 0.0214601 |
| 5 | [0.127, 0.139) | 0.133028 | 0.0088771 | 0.0109435 | 0.0208206 | 0.0393504 | 0.0481357 | 0.0236802 |
| 6 | [0.139, 0.151) | 0.144977 | 0.0097898 | 0.0118208 | 0.0225534 | 0.0439163 | 0.0538868 | 0.0259512 |
| 7 | [0.151, 0.163) | 0.156874 | 0.0101678 | 0.0126009 | 0.0248914 | 0.0500528 | 0.0613696 | 0.0288113 |
| 8 | [0.163, 0.176) | 0.169306 | 0.0113233 | 0.0138144 | 0.0274455 | 0.0552421 | 0.0656847 | 0.0317599 |
| 9 | [0.176, 0.188) | 0.181783 | 0.0120341 | 0.0147276 | 0.0297646 | 0.0587297 | 0.0700824 | 0.0339487 |
| 10 | [0.188, 0.200) | 0.193745 | 0.0124796 | 0.0157621 | 0.0317564 | 0.0613300 | 0.0720410 | 0.0356610 |
| 11 | [0.200, 0.212) | 0.205758 | 0.0137977 | 0.0167981 | 0.0329443 | 0.0622977 | 0.0721443 | 0.0368687 |
| 12 | [0.212, 0.225) | 0.218204 | 0.0140878 | 0.0175574 | 0.0347152 | 0.0652521 | 0.0774105 | 0.0386492 |
| 13 | [0.225, 0.237) | 0.230723 | 0.0140705 | 0.0177333 | 0.0369638 | 0.0672822 | 0.0764555 | 0.0402684 |

| 14 | [0.237, 0.249) | 0.242818 | 0.0146810 | 0.0186397 | 0.0381558 | 0.0692375 | 0.0786225 | 0.0413928 |
|---|---|---|---|---|---|---|---|---|
| 15 | [0.249, ∞) | 0.279499 | 0.0168309 | 0.0207854 | 0.0401272 | 0.0721349 | 0.0829652 | 0.0439267 |
| $i$ | VRCM$_i$ for CM($LTI_2$) | Mean CM($LTI_2$) in VRCM$_i$ | $p$–quantiles of MAEs among **w** and **w***(LLSM) | | | | | Average MAEs among **w** and **w***(LLSM) |
| | | | $p$=0.05 | $p$=0.1 | $p$=0.5 | $p$=0.9 | $p$=0.95 | |
| 1 | [0, 0.07975) | 0.061626 | 0.00329141 | 0.0040063 | 0.0079292 | 0.0153184 | 0.017980 | 0.0089902 |
| 2 | [0.07975, 0.09) | 0.085354 | 0.00558449 | 0.0066781 | 0.0124836 | 0.0217916 | 0.0254877 | 0.0136394 |
| 3 | [0.09, 0.10) | 0.095128 | 0.00622084 | 0.0074651 | 0.0136346 | 0.0241580 | 0.0288301 | 0.0151410 |
| 4 | [0.10, 0.11) | 0.105046 | 0.00677146 | 0.0081844 | 0.0150571 | 0.0277432 | 0.0343089 | 0.0170348 |
| 5 | [0.11, 0.12) | 0.114884 | 0.00728075 | 0.0089529 | 0.0164708 | 0.0329745 | 0.0408782 | 0.0192642 |
| 6 | [0.12, 0.13) | 0.124902 | 0.00792417 | 0.0097471 | 0.0185168 | 0.0378364 | 0.0464765 | 0.0217170 |
| 7 | [0.13, 0.14) | 0.134949 | 0.00851189 | 0.0104389 | 0.0202075 | 0.0415434 | 0.0507830 | 0.0236614 |
| 8 | [0.14, 0.15) | 0.144883 | 0.00952136 | 0.0115606 | 0.0224145 | 0.0446314 | 0.0531641 | 0.0257116 |
| 9 | [0.15, 0.161) | 0.155416 | 0.0101888 | 0.0121602 | 0.0241178 | 0.0465694 | 0.0553538 | 0.0275101 |
| 10 | [0.161, 0.171) | 0.165845 | 0.0110535 | 0.0132394 | 0.0261677 | 0.0499157 | 0.0583309 | 0.0293786 |
| 11 | [0.171, 0.181) | 0.175874 | 0.0116123 | 0.0139639 | 0.0273006 | 0.0515428 | 0.0596329 | 0.0304575 |
| 12 | [0.181, 0.191) | 0.185981 | 0.0121824 | 0.0150547 | 0.0293308 | 0.0532065 | 0.0613544 | 0.0320030 |
| 13 | [0.191, 0.201) | 0.195819 | 0.0122294 | 0.0152015 | 0.0299135 | 0.0553010 | 0.0642011 | 0.0330142 |
| 14 | [0.201, 0.211) | 0.205937 | 0.0132402 | 0.0164008 | 0.0321805 | 0.0552598 | 0.0636846 | 0.0343310 |
| 15 | [0.211, ∞) | 0.235348 | 0.0147413 | 0.0179580 | 0.0321805 | 0.0586515 | 0.0682411 | 0.0363445 |
| $i$ | VRCM$_i$ for CM($LTI_2$) | Mean CM($LTI_2$) in VRCM$_i$ | $p$–quantiles of MAEs among **w** and **w***(LLSM) | | | | | Average MAEs among **w** and **w***(LLSM) |
| | | | $p$=0.05 | $p$=0.1 | $p$=0.5 | $p$=0.9 | $p$=0.95 | |
| 1 | [0, 0.06861) | 0.056493 | 0.0029930 | 0.0036359 | 0.0071723 | 0.0129253 | 0.0151100 | 0.0079193 |
| 2 | [0.06861, 0.078) | 0.073616 | 0.0047668 | 0.0056647 | 0.0098201 | 0.0165545 | 0.0197820 | 0.0107659 |
| 3 | [0.078, 0.087) | 0.082558 | 0.0051148 | 0.00615425 | 0.0108293 | 0.0189720 | 0.0232764 | 0.0121106 |
| 4 | [0.087, 0.095) | 0.090957 | 0.0054815 | 0.0067644 | 0.0120701 | 0.0230822 | 0.0289636 | 0.0139455 |
| 5 | [0.095, 0.104) | 0.0995085 | 0.0062208 | 0.0074045 | 0.0134360 | 0.0267488 | 0.0338094 | 0.0157422 |
| 6 | [0.104, 0.113) | 0.108507 | 0.0065308 | 0.0079148 | 0.0148310 | 0.0307300 | 0.0379708 | 0.0175495 |
| 7 | [0.113, 0.122) | 0.117503 | 0.0073636 | 0.0087983 | 0.0166204 | 0.0342287 | 0.0402093 | 0.0192815 |
| 8 | [0.122, 0.131) | 0.126447 | 0.0077367 | 0.00920785 | 0.0182781 | 0.0366835 | 0.0432579 | 0.0209778 |
| 9 | [0.131, 0.140) | 0.135467 | 0.0081883 | 0.0099817 | 0.0200944 | 0.0391024 | 0.0463982 | 0.0227669 |
| 10 | [0.140, 0.149) | 0.144406 | 0.00893715 | 0.0109052 | 0.0215995 | 0.0404294 | 0.0465999 | 0.0240918 |
| 11 | [0.149, 0.158) | 0.153395 | 0.0096365 | 0.0118788 | 0.0228543 | 0.0420224 | 0.0488816 | 0.0252208 |
| 12 | [0.158, 0.167) | 0.162379 | 0.0105213 | 0.0128739 | 0.0250637 | 0.0441591 | 0.0509963 | 0.0270496 |
| 13 | [0.167, 0.176) | 0.171319 | 0.0109917 | 0.0133182 | 0.0253654 | 0.0446033 | 0.0525163 | 0.0275815 |
| 14 | [0.176, 0.185) | 0.180246 | 0.0120041 | 0.0144395 | 0.0266159 | 0.0464516 | 0.0529339 | 0.0289197 |
| 15 | [0.185, ∞) | 0.205854 | 0.0127740 | 0.0155662 | 0.0283804 | 0.0479564 | 0.0549310 | 0.0304352 |
| $i$ | VRCM$_i$ for CM($LTI_2$) | Mean CM($LTI_2$) in VRCM$_i$ | $p$–quantiles of MAEs among **w** and **w***(LLSM) | | | | | Average MAEs among **w** and **w***(LLSM) |
| | | | $p$=0.05 | $p$=0.1 | $p$=0.5 | $p$=0.9 | $p$=0.95 | |
| 1 | [0, 0.059795) | 0.051197 | 0.0026372 | 0.0031677 | 0.0061278 | 0.0107141 | 0.0122502 | 0.0066588 |
| 2 | [0.05979, 0.068) | 0.064092 | 0.0040166 | 0.0047872 | 0.0079722 | 0.0133870 | 0.0158901 | 0.0087678 |
| 3 | [0.068, 0.076) | 0.072019 | 0.0044127 | 0.0052033 | 0.0089180 | 0.0154595 | 0.0189767 | 0.0099818 |
| 4 | [0.076, 0.085) | 0.080495 | 0.0046625 | 0.0055923 | 0.0098746 | 0.0183837 | 0.0234965 | 0.0114040 |
| 5 | [0.085, 0.093) | 0.089047 | 0.0052813 | 0.0062378 | 0.0110158 | 0.0221043 | 0.0279174 | 0.0129826 |
| 6 | [0.093, 0.101) | 0.097017 | 0.0056575 | 0.0067669 | 0.0124636 | 0.0261396 | 0.0326188 | 0.0147615 |
| 7 | [0.101, 0.109) | 0.105051 | 0.0061505 | 0.0074036 | 0.0138920 | 0.0290254 | 0.0358010 | 0.0164984 |
| 8 | [0.109, 0.118) | 0.113488 | 0.0066692 | 0.0079686 | 0.0153474 | 0.0308319 | 0.0365922 | 0.0177312 |
| 9 | [0.118, 0.126) | 0.122009 | 0.0073133 | 0.0087907 | 0.0171076 | 0.0330852 | 0.0388189 | 0.0193438 |
| 10 | [0.126, 0.134) | 0.129857 | 0.0076181 | 0.0092912 | 0.0186416 | 0.0343317 | 0.0401595 | 0.0204982 |
| 11 | [0.134, 0.142) | 0.137970 | 0.0083801 | 0.0102174 | 0.0199779 | 0.0355818 | 0.0416818 | 0.0216939 |
| 12 | [0.142, 0.151) | 0.146298 | 0.0091112 | 0.0107807 | 0.0212040 | 0.0376109 | 0.0430078 | 0.0229256 |
| 13 | [0.151, 0.159) | 0.154883 | 0.0097330 | 0.0118942 | 0.0219635 | 0.0378245 | 0.0435528 | 0.0237785 |
| 14 | [0.159, 0.167) | 0.162793 | 0.0102563 | 0.0125995 | 0.0228089 | 0.0390630 | 0.0443591 | 0.0244409 |
| 15 | [0.167, ∞) | 0.184864 | 0.0115601 | 0.0138072 | 0.0242891 | 0.0403012 | 0.046879 | 0.0259996 |

**Table A2 – Performance of CM(*LTI₂*) index under the action of LUA as the PP**. Statistical characteristics of the MAEs distribution in relation to various levels of CM(*LTI₂*) within a given VRCM$_i$ for $i$=1,…,15. The results are based on 10,000 perturbed random reciprocal PCMs (geometric and Saaty's scales applied fifty-fifty), and were generated on the basis of **SA|2|** as the simulation algorithm. The table contains results for $n \in \{5, 6, 7, 8, 9\}$, presented consecutively.

| $i$ | VRCM$_i$ for CM(*LTI₂*) | Mean CM(*LTI₂*) in VRCM$_i$ | *p*–quantiles of MAEs among **w** and **w***(LUA) | | | | | Average MAEs among **w** and **w***(LUA) |
|---|---|---|---|---|---|---|---|---|
| | | | *p*=0.05 | *p*=0.1 | *p*=0.5 | *p*=0.9 | *p*=0.95 | |
| 1 | [0, 0.08867) | 0.057344 | 0.0040500 | 0.0051112 | 0.0109956 | 0.0223145 | 0.0281738 | 0.0129222 |
| 2 | [0.08867, 0.106) | 0.097631 | 0.0051370 | 0.0066813 | 0.0149743 | 0.0314682 | 0.0402326 | 0.0179091 |
| 3 | [0.106, 0.123) | 0.115154 | 0.0062273 | 0.0080033 | 0.0178583 | 0.0404214 | 0.0493824 | 0.0215433 |
| 4 | [0.123, 0.140) | 0.132429 | 0.0077722 | 0.0103867 | 0.0235191 | 0.0443250 | 0.0533440 | 0.0260091 |
| 5 | [0.140, 0.158) | 0.149841 | 0.0100660 | 0.0130848 | 0.0264063 | 0.0492151 | 0.0598150 | 0.0293785 |
| 6 | [0.158, 0.175) | 0.166943 | 0.0122130 | 0.0152940 | 0.0305507 | 0.0567287 | 0.0669818 | 0.0339792 |
| 7 | [0.175, 0.192) | 0.183544 | 0.0134146 | 0.0168104 | 0.0341529 | 0.0622582 | 0.0730947 | 0.0376190 |
| 8 | [0.192, 0.209) | 0.200556 | 0.0144681 | 0.0180775 | 0.0371079 | 0.0664060 | 0.0801886 | 0.0405209 |
| 9 | [0.209, 0.227) | 0.217798 | 0.0152484 | 0.0195489 | 0.0387389 | 0.0726297 | 0.0886177 | 0.0432569 |
| 10 | [0.227, 0.244) | 0.235136 | 0.0161576 | 0.0201625 | 0.0403835 | 0.0771441 | 0.0945720 | 0.0454089 |
| 11 | [0.244, 0.261) | 0.252143 | 0.0164634 | 0.0205743 | 0.0428687 | 0.0812496 | 0.0997771 | 0.0479053 |
| 12 | [0.261, 0.278) | 0.269128 | 0.0174125 | 0.0217309 | 0.0445472 | 0.0844031 | 0.1015070 | 0.0498806 |
| 13 | [0.278, 0.296) | 0.286560 | 0.0184856 | 0.0235664 | 0.0474587 | 0.0907092 | 0.1046180 | 0.0527022 |
| 14 | [0.296, 0.313) | 0.304366 | 0.0176996 | 0.0228077 | 0.0479535 | 0.0900992 | 0.1047390 | 0.0532388 |
| 15 | [0.313, ∞) | 0.354236 | 0.0192203 | 0.0244908 | 0.0503011 | 0.0929620 | 0.1098880 | 0.0556579 |
| $i$ | VRCM$_i$ for CM(*LTI₂*) | Mean CM(*LTI₂*) in VRCM$_i$ | *p*–quantiles of MAEs among **w** and **w***(LUA) | | | | | Average MAEs among **w** and **w***(LUA) |
| | | | *p*=0.05 | *p*=0.1 | *p*=0.5 | *p*=0.9 | *p*=0.95 | |
| 1 | [0, 0.09033) | 0.063185 | 0.0035880 | 0.0044267 | 0.0089867 | 0.0179120 | 0.0222610 | 0.0103740 |
| 2 | [0.09033, 0.103) | 0.097359 | 0.0063423 | 0.0078999 | 0.0155573 | 0.0273932 | 0.0319008 | 0.0168600 |
| 3 | [0.103, 0.115) | 0.109410 | 0.00722965 | 0.0091969 | 0.0177403 | 0.0325171 | 0.0386338 | 0.0196173 |
| 4 | [0.115, 0.127) | 0.121343 | 0.0084960 | 0.0107027 | 0.0210117 | 0.0381138 | 0.0455231 | 0.0233632 |
| 5 | [0.127, 0.139) | 0.133108 | 0.0094898 | 0.0116495 | 0.0229152 | 0.0420076 | 0.0524375 | 0.0257261 |
| 6 | [0.139, 0.152) | 0.145538 | 0.0108994 | 0.0132421 | 0.0253036 | 0.0481306 | 0.0607595 | 0.0287714 |
| 7 | [0.152, 0.164) | 0.157728 | 0.0114276 | 0.0139558 | 0.0271455 | 0.0539605 | 0.0656762 | 0.0310679 |
| 8 | [0.164, 0.176) | 0.169681 | 0.0121504 | 0.0150640 | 0.0292961 | 0.0575169 | 0.0691811 | 0.0336092 |
| 9 | [0.176, 0.189) | 0.182266 | 0.0128272 | 0.0159433 | 0.0313538 | 0.0612481 | 0.0726740 | 0.0356428 |
| 10 | [0.189, 0.201) | 0.194801 | 0.0138039 | 0.0173398 | 0.0328835 | 0.0629099 | 0.0736509 | 0.0370798 |
| 11 | [0.201, 0.213) | 0.206815 | 0.0140270 | 0.0173321 | 0.0347352 | 0.0651391 | 0.0772600 | 0.0387081 |
| 12 | [0.213, 0.225) | 0.218752 | 0.0145777 | 0.0184396 | 0.0366233 | 0.0669927 | 0.0785993 | 0.0400494 |
| 13 | [0.225, 0.238) | 0.231199 | 0.0152711 | 0.0185568 | 0.0383407 | 0.0700653 | 0.0825784 | 0.0420203 |
| 14 | [0.238, 0.250) | 0.243637 | 0.0158614 | 0.0199098 | 0.0378087 | 0.0715807 | 0.0841889 | 0.0422926 |
| 15 | [0.250, ∞) | 0.280974 | 0.0171267 | 0.0210891 | 0.0412814 | 0.0738420 | 0.0859174 | 0.0450109 |
| $i$ | VRCM$_i$ for CM(*LTI₂*) | Mean CM(*LTI₂*) in VRCM$_i$ | *p*–quantiles of MAEs among **w** and **w***(LUA) | | | | | Average MAEs among **w** and **w***(LUA) |
| | | | *p*=0.05 | *p*=0.1 | *p*=0.5 | *p*=0.9 | *p*=0.95 | |
| 1 | [0, 0.07955) | 0.061285 | 0.0032254 | 0.0039749 | 0.0080564 | 0.0157394 | 0.0187853 | 0.0091669 |
| 2 | [0.07955, 0.09) | 0.085237 | 0.0055756 | 0.00703635 | 0.0133354 | 0.0239340 | 0.0276372 | 0.0145973 |
| 3 | [0.09, 0.10) | 0.095217 | 0.0067076 | 0.00822896 | 0.0151476 | 0.0267136 | 0.0313365 | 0.0168334 |
| 4 | [0.10, 0.11) | 0.105055 | 0.0071842 | 0.0087319 | 0.0167347 | 0.0300692 | 0.0372910 | 0.0187971 |
| 5 | [0.110, 0.120) | 0.114948 | 0.0078851 | 0.0096590 | 0.0184274 | 0.0350956 | 0.0451092 | 0.0211768 |
| 6 | [0.120, 0.130) | 0.124976 | 0.00841585 | 0.0104132 | 0.0202074 | 0.0399982 | 0.0487099 | 0.0233008 |
| 7 | [0.130, 0.140) | 0.134961 | 0.0094352 | 0.0114042 | 0.0217320 | 0.0442261 | 0.0533524 | 0.0258727 |
| 8 | [0.140, 0.150) | 0.144944 | 0.0097585 | 0.0120015 | 0.0234560 | 0.0471241 | 0.0565387 | 0.0269760 |
| 9 | [0.150, 0.160) | 0.154896 | 0.0105670 | 0.0127710 | 0.0253768 | 0.0489510 | 0.0573372 | 0.0286380 |
| 10 | [0.160, 0.171) | 0.165237 | 0.0113250 | 0.0138414 | 0.0270817 | 0.0499840 | 0.0598609 | 0.0301303 |
| 11 | [0.171, 0.181) | 0.175843 | 0.0120019 | 0.0146386 | 0.0287811 | 0.0534214 | 0.0617005 | 0.0317433 |
| 12 | [0.181, 0.191) | 0.185759 | 0.0126595 | 0.0153651 | 0.0298787 | 0.0548531 | 0.0642158 | 0.0329712 |
| 13 | [0.191, 0.201) | 0.195725 | 0.0128283 | 0.0154763 | 0.0313703 | 0.0562006 | 0.0639533 | 0.0339822 |
| 14 | [0.201, 0.211) | 0.205744 | 0.0139552 | 0.0170698 | 0.0326421 | 0.0578949 | 0.0674790 | 0.0354460 |
| 15 | [0.211, ∞) | 0.235674 | 0.0149229 | 0.0180402 | 0.0341664 | 0.0601557 | 0.0694405 | 0.0371047 |
| $i$ | VRCM$_i$ for CM(*LTI₂*) | Mean CM(*LTI₂*) in VRCM$_i$ | *p*–quantiles of MAEs among **w** and **w***(LUA) | | | | | Average MAEs among **w** and **w***(LUA) |
| | | | *p*=0.05 | *p*=0.1 | *p*=0.5 | *p*=0.9 | *p*=0.95 | |
| 1 | [0, 0.0688) | 0.056836 | 0.0030787 | 0.0036205 | 0.0073926 | 0.0145693 | 0.0176789 | 0.0084662 |
| 2 | [0.0688, 0.078) | 0.073666 | 0.0051105 | 0.0061309 | 0.0109223 | 0.0189602 | 0.0222010 | 0.0121584 |
| 3 | [0.078, 0.087) | 0.082561 | 0.0055045 | 0.0066116 | 0.0121236 | 0.0213335 | 0.0264824 | 0.0136680 |

| i | VRCM$_i$ for CM($LTI_2$) | Mean CM($LTI_2$) in VRCM$_i$ | p–quantiles of MAEs among **w** and **w*** (LUA) | | | | | Average MAEs among **w** and **w*** (LUA) |
|---|---|---|---|---|---|---|---|---|
| | | | p=0.05 | p=0.1 | p=0.5 | p=0.9 | p=0.95 | |
| 4 | [0.087, 0.096) | 0.091556 | 0.0059645 | 0.0072229 | 0.0133996 | 0.0253087 | 0.0332026 | 0.0154277 |
| 5 | [0.096, 0.105) | 0.100528 | 0.0063149 | 0.0077633 | 0.0145624 | 0.0296739 | 0.0375550 | 0.0170392 |
| 6 | [0.105, 0.114) | 0.109496 | 0.0069818 | 0.0085509 | 0.0161950 | 0.0334936 | 0.0413081 | 0.0189646 |
| 7 | [0.114, 0.123) | 0.118520 | 0.0075613 | 0.0090468 | 0.0175431 | 0.0357544 | 0.0434245 | 0.0204242 |
| 8 | [0.123, 0.132) | 0.127416 | 0.0079922 | 0.0097597 | 0.0192472 | 0.0395417 | 0.0473757 | 0.0223206 |
| 9 | [0.132, 0.141) | 0.136453 | 0.0086852 | 0.0105264 | 0.0209858 | 0.0407715 | 0.0482480 | 0.0238363 |
| 10 | [0.141, 0.150) | 0.145395 | 0.0097858 | 0.0118143 | 0.0228944 | 0.0419213 | 0.0491251 | 0.0254135 |
| 11 | [0.150, 0.159) | 0.154468 | 0.0100095 | 0.0125284 | 0.0241689 | 0.0438311 | 0.0507983 | 0.0265045 |
| 12 | [0.159, 0.168) | 0.163355 | 0.0106847 | 0.0132036 | 0.0259776 | 0.0454986 | 0.0526910 | 0.0279660 |
| 13 | [0.168, 0.177) | 0.172363 | 0.0111935 | 0.0138710 | 0.0271393 | 0.0471614 | 0.0543407 | 0.0291270 |
| 14 | [0.177, 0.186) | 0.181389 | 0.0124251 | 0.0149686 | 0.0273290 | 0.0474666 | 0.0540064 | 0.0294350 |
| 15 | [0.186, ∞) | 0.206400 | 0.0133439 | 0.0162347 | 0.0294684 | 0.0496426 | 0.0567018 | 0.0315851 |
| i | VRCM$_i$ for CM($LTI_2$) | Mean CM($LTI_2$) in VRCM$_i$ | p–quantiles of MAEs among **w** and **w*** (LUA) | | | | | Average MAEs among **w** and **w*** (LUA) |
| | | | p=0.05 | p=0.1 | p=0.5 | p=0.9 | p=0.95 | |
| 1 | [0, 0.05999) | 0.051571 | 0.0027899 | 0.0033529 | 0.0066145 | 0.0129846 | 0.0152304 | 0.00757575 |
| 2 | [0.05999, 0.068) | 0.064108 | 0.00425595 | 0.0049956 | 0.0089923 | 0.0154707 | 0.0180262 | 0.0099654 |
| 3 | [0.068, 0.076) | 0.0720425 | 0.0045459 | 0.0053806 | 0.0096431 | 0.0170663 | 0.0215704 | 0.0110540 |
| 4 | [0.076, 0.085) | 0.080557 | 0.00484755 | 0.0059128 | 0.0107496 | 0.0200923 | 0.0264333 | 0.0124749 |
| 5 | [0.085, 0.093) | 0.089092 | 0.0053555 | 0.0064602 | 0.0118465 | 0.0241501 | 0.0307256 | 0.0140711 |
| 6 | [0.093, 0.101) | 0.096974 | 0.0059146 | 0.0070858 | 0.0130955 | 0.0270450 | 0.0340523 | 0.0154649 |
| 7 | [0.101, 0.109) | 0.105011 | 0.0064092 | 0.0077271 | 0.0143894 | 0.0307524 | 0.0372307 | 0.0172088 |
| 8 | [0.109, 0.118) | 0.113412 | 0.0070543 | 0.0083974 | 0.0158975 | 0.0323549 | 0.0386885 | 0.0184995 |
| 9 | [0.118, 0.126) | 0.121966 | 0.0076984 | 0.0092627 | 0.0180989 | 0.0344896 | 0.0405523 | 0.0204213 |
| 10 | [0.126, 0.134) | 0.129881 | 0.0081201 | 0.0097966 | 0.0194020 | 0.0364334 | 0.0431290 | 0.0216106 |
| 11 | [0.134, 0.142) | 0.137905 | 0.0088861 | 0.0106243 | 0.0206471 | 0.0371653 | 0.0437476 | 0.0225820 |
| 12 | [0.142, 0.151) | 0.146345 | 0.0094580 | 0.0114928 | 0.0216677 | 0.0380785 | 0.0443759 | 0.0235540 |
| 13 | [0.151, 0.159) | 0.154771 | 0.0101348 | 0.0124254 | 0.0231652 | 0.0396143 | 0.0456328 | 0.0248599 |
| 14 | [0.159, 0.167) | 0.162795 | 0.0108284 | 0.0129963 | 0.0234617 | 0.0396838 | 0.0458155 | 0.0250440 |
| 15 | [0.167, ∞) | 0.184832 | 0.0122591 | 0.0145491 | 0.0255388 | 0.0426019 | 0.0490175 | 0.0273476 |